\documentclass[10pt,twocolumn,letterpaper]{article}

\usepackage[pagenumbers]{cvpr} % To force page numbers, e.g. for an arXiv version

\usepackage{graphicx}
\usepackage{amsmath}
\usepackage{amssymb}
\usepackage{booktabs}
\usepackage[accsupp]{axessibility}

\usepackage[normalem]{ulem}
\newcommand{\revadd}[1]{\textcolor{black}{#1}}

\newcommand{\vect}[1]{\textbf{#1}}

\usepackage[pagebackref,breaklinks,colorlinks]{hyperref}

\usepackage[capitalize]{cleveref}
\crefname{section}{Sec.}{Secs.}
\Crefname{section}{Section}{Sections}
\Crefname{table}{Table}{Tables}
\crefname{table}{Tab.}{Tabs.}

\def\cvprPaperID{11462} % *** Enter the CVPR Paper ID here
\def\confName{CVPR}
\def\confYear{2023}

\begin{document}

\title{Implicit View-Time Interpolation of Stereo Videos using Multi-Plane Disparities and Non-Uniform Coordinates}

\author{Avinash Paliwal$^{1}$\hspace{0.1in} 
        Andrii Tsarov$^{2}$\hspace{0.1in} 
        Nima Khademi Kalantari$^{1}$\hspace{0.1in}\\
$^1$Texas A\&M University\hspace{0.2in}
$^2$Leia Inc.\\
{\tt\small \{avinashpaliwal,nimak\}@tamu.edu, andrii.tsarov@leiainc.com}\\
% \vspace{-2mm}
}
\maketitle

%%%%%%%%% ABSTRACT
\begin{abstract}
   In this paper, we propose an approach for view-time interpolation of stereo videos. Specifically, we build upon X-Fields that approximates an interpolatable mapping between the input coordinates and 2D RGB images using a convolutional decoder. Our main contribution is to analyze and identify the sources of the problems with using X-Fields in our application and propose novel techniques to overcome these challenges. Specifically, we observe that X-Fields struggles to implicitly interpolate the disparities for large baseline cameras. Therefore, we propose multi-plane disparities to reduce the spatial distance of the objects in the stereo views. Moreover, we propose non-uniform time coordinates to handle the non-linear and sudden motion spikes in videos. We additionally introduce several simple, but important, improvements over X-Fields. We demonstrate that our approach is able to produce better results than the state of the art, while running in near real-time rates and having low memory and storage costs.
\end{abstract}

%%%%%%%%% BODY TEXT
\vspace{-2mm}

\section{Introduction}
\label{sec:Introduction}
As the virtual reality (VR) and light field displays (e.g., Lume Pad~\cite{leia}) become widespread, there is a growing need for capturing the appropriate content for these devices to provide an immersive virtual experience for the users. This necessitates capturing a scene from different views and at high frame rates. While this can be done using specialized hardware~\cite{broxton2020immersive}, such setups are usually bulky and expensive, and thus not suitable for an average user. To make the content capture widespread, we should focus on standard capturing devices like cellphone cameras. These devices, however, are typically equipped with only two cameras and are often not able to capture two videos at high frame rates. This necessitates interpolating across time and view to reconstruct high frame rate videos from dense views.

For a system to be practical and can be deployed on display devices with limited storage, memory, and computational capability, it should have a few properties: \textbf{1)} while a reasonable amount of post-processing can be done on a server, the approach should be able to generate results in real-time, \textbf{2)} it should not have a significant storage overhead on top of the input stereo video, and \textbf{3)} it should have a low memory cost. 

Unfortunately, most existing approaches violate one or more criteria. For example, while the approaches based on multi-plane images (MPI)~\cite{Zhou2018stereo,Srinivasan2019pushing,Wizadwongsa2021nex} can render novel views in real-time, they require storing the estimated MPIs for each frame (tens of megabytes per frame) and are additionally memory intensive. Moreover, to perform both view and time interpolations, these approaches need to be augmented with a video interpolation method which further adds to their memory and computational cost. The more recent approaches based on neural radiance fields (NeRF)~\cite{Mildenhall2020nerf} can perform both view and time interpolations~\cite{Du2021nerflow,Li2020neural}. These methods encode the radiance field of a scene into a small network, and thus have a small storage and memory cost. However, they usually take a few seconds to render each novel view. Additionally, they require the cameras to be calibrated, and thus have difficulty handling general videos. 

In this paper, we build upon X-Fields~\cite{Bemana2020xfields} that optimizes a coordinate-based network to learn an interpolatable implicit mapping between the input coordinates X (view or time) and the observed images. Once the optimization for a specific scene is performed, the network can be used to generate an image given any X coordinate. This approach satisfies all the properties, as the rendering is real-time (criterion 1) and the scene is encoded into a small network (criteria 2 and 3). Despite that, X-Fields struggles to produce reasonable results for the specific problem of view-time interpolation of stereo videos.

Our main contribution is to analyze X-Fields, identify the sources of the problems, and propose approaches that address these shortfalls. Specifically, we identify two major problems with X-Fields for stereo video interpolation. Our first observation is that X-Fields struggles to interpolate the disparities for cameras with large baselines. Second, we observe that linear motion in the input videos is critical for X-Fields optimization to work, but non-linear motions are common in natural videos. 

We address the first issue by proposing multi-plane disparities to reduce the spatial distance of the objects in the scene. Our approach makes it substantially easier for the network to interpolate the disparities, as the left and right disparities in different planes become closer to each other spatially. Moreover, we propose to address the second problem through a novel non-uniform time coordinate encoding method. We demonstrate that the implicit network is able to find a proper mapping between these non-uniform coordinates and the observations and has a significantly better interpolation capability.

Additionally, we propose a series of simple, but important, improvements such as additional regularization losses, learned blending, and positional encoding. Our approach has low memory and storage costs, and is able to reconstruct novel images in near real-time rates. Through extensive experiments we demonstrate that our method is significantly better than the state-of-the-art approaches. Code and supplementary materials are available on our project website at \href{https://people.engr.tamu.edu/nimak/Papers/CVPR23StereoVideo/index.html}{\small https://people.engr.tamu.edu/nimak/Papers/CVPR23StereoVideo}.%{\small~\url{https://people.engr.tamu.edu/nimak/Papers/CVPR23_StereoVideo}}.

\section{Related Work}

In this section, we briefly review the relevant view, time, and view-time synthesis methods.

%-------------------------------------------------------------------------
\subsection{View Synthesis}
Novel view synthesis is a widely studied problem. A classical solution is to first reconstruct the 3D geometry of the scene, e.g., point clouds, and then render novel views based on the geometry ~\cite{Debevec1996modeling,Beuhler2001unstructured,Chaurasia2013depth,Hedman2017casual,Klose2015sampling,Snavely2006photo,Hedman2018instant,Penner2017soft}. With the rise of deep learning, several approaches propose to handle this application using a neural network. For example, Flynn et al.~\cite{Flynn_2016} proposes a network based on plane sweep volumes, while Kalantari et al.~\cite{Kalantari_2016} estimates the disparity and blending weights using two sequential networks. Zhou et al.~\cite{Zhou2018stereo} introduce multi-plane images (MPI), a flexible scene representation which is suitable for view synthesis. Several approaches~\cite{Wizadwongsa2021nex, Srinivasan2019pushing, mildenhall2019local,Flynn2019deepview,lin2021deep,Li2020synthesizing} propose various ways to extend this idea. The major problem with these approaches is significant storage and memory costs. Following the introduction of neural radiance field (NeRF) by Mildenhall et al.~\cite{Mildenhall2020nerf}, a large number of approaches~\cite{Peng2021neural,Srinivasan2021nerv,Hedman2021baking,Jain2021putting,kangle2021depth,park2021nerfies,park2021hypernerf,Gao2021Dynamic} based on this idea have been proposed. While these approaches are powerful, they are typically slow to render novel images.

%-------------------------------------------------------------------------
\subsection{Time Interpolation}
Most video interpolation approaches rely on optical flow to warp images and synthesize the interpolated frames. Recent state-of-the-art methods rely on deep learning for flow computation and image synthesis~\cite{Jiang2018super,Bao2019depth,Niklaus2018context,Niklaus2020softmax,park2020bmbc}. Several methods~\cite{kalluri2021flavr,choi2020channel} propose to directly generate the interpolated images without explicitly estimating a flow. Finally, a few methods~\cite{Choi2020scene,Reda2019unsupervised} adapt a network on the test example at hand by fine tuning it through the typical appearance loss.
%-------------------------------------------------------------------------
\subsection{View-Time Synthesis}
Several recent approaches~\cite{Tretschk2021nonrigid,Xian2021space,Pumarola2021dnerf,Du2021nerflow,Li2020neural} extend NeRF to handle the additional time dimension and work on dynamic scenes. In addition to the common shortfalls of the NeRF-based approaches, these methods can only handle limited types of videos, as they require the cameras to be calibrated. Different from these approaches, Bemana et al.~\cite{Bemana2020xfields} propose X-Fields, a lightweight network capable of real-time view-time (as well as additional dimensions like light) interpolation. This approach essentially learns a per-scene mapping between the coordinates and 2D RGB images. We build upon X-Fields, but propose key ideas to significantly improve its performance on the problem of view-time interpolation of stereo videos.

\section{Background}

Given a set of images captured with different modalities (e.g., view and time), X-Fields~\cite{Bemana2020xfields} poses the problem as approximating the mapping between the input coordinates X and the corresponding images through a small coordinate-based neural network. To do so, X-Fields optimizes the network using the following objective:

% \vspace{-0.1in}
\begin{equation}
    \label{eq:xfields_loss}
    \theta^* = \arg \min_\theta \sum_{i = 1}^N \Vert f_\theta(\vect{y}_i) - f(\vect{y}_i) \Vert_1
\end{equation}
% \vspace{-0.1in}

\noindent where $f(\vect{y}_i)$ is the observed image at coordinate $\vect{y}_i$ and $f_\theta(\vect{y}_i)$ is the approximated image using the network with weights $\theta$. The approximated mapping function $f_\theta$ can then be used to reconstruct an image from any novel coordinate $\vect{x}$ (in the convex hull of the observed coordinates).

Instead of directly estimating the 2D RGB images using the network, X-Fields reconstructs the novel image by warping and combining the observed images in the neighborhood of the coordinate of interest. This is done by first estimating the Jacobian of the flows at each pixel. This Jacobian describes how a pixel in the coordinate of interest moves if, for example, the time coordinate changes. The Jacobian which is estimated by the network is defined as:

% \vspace{-0.1in}
\begin{equation}
    \label{eq:jacobian}
    g_\theta(\vect{x})[\vect{p}] = J(\vect{x})[\vect{p}] = \frac{\partial \vect{p}(\vect{x}) }{\partial \vect{x}},
\end{equation}
% \vspace{-0.1in}

\noindent where $g_\theta$ is the network and $J(\vect{x})[\vect{p}]$ denotes the Jacobian at pixel $\vect{p}$ of coordinate $\vect{x}$. Using this Jacobian, the flow from pixel $\vect{p}(\vect{x})$ to the corresponding pixel $\vect{q}$ in coordinate $\vect{y}$ can be obtained by:

% \vspace{-0.1in}
\begin{equation}
    \label{eq:flow}
    \vect{F}_\vect{y}(\vect{x})[\vect{p}] = \vect{p} + (\vect{y} - \vect{x}) J(\vect{x})[\vect{p}],
\end{equation}
\vspace{-0.1in}

\noindent where $\vect{F}_\vect{y}(\vect{x})$ refers to the flow from the image at coordinate $\vect{x}$ to the one at $\vect{y}$. Using these flows, the observed neighboring images are warped to the coordinate of interest and are combined using a weight map, derived from the forward-backward flow consistency check.

\section{Algorithm}
\label{sec:Algorithm}

Given a stereo video with $N$ frames, our goal is to reconstruct images at novel views and  times. We build upon X-Fields and attempt to learn an implicit mapping from the view-time coordinates $\vect{x}$ to the corresponding 2D RGB images $f(\vect{x})$. In our specific problem, the coordinates are 2 dimensional $\vect{x} = (u, t)$, where $u$ is one dimensional view between and including the two views, and $t$ denotes the one dimensional time coordinate.

We begin by discussing the view synthesis and time interpolation separately in Secs.~\ref{ssec:StereoView}~and~\ref{ssec:TimeInterpolation}, respectively. We then discuss the view-time interpolation in Sec.~\ref{ssec:view_time}.

\begin{figure}
\includegraphics[width=\linewidth]{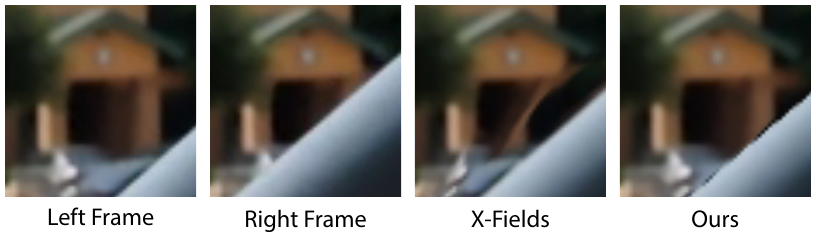}
\vspace{-0.3in}
\caption{We show the images from the left and right views on the left. The interpolated middle images using only appearance loss (X-Fields) and our results are shown on the right. The stretching artifacts around the depth discontinuities because of the poorly estimated Jacobians can be observed in X-Fields results.}
\label{fig:blendinset}
\vspace{-0.3in}
\end{figure}

\subsection{View Synthesis}
\label{ssec:StereoView}

Given a pair of stereo images at a specific frame, our goal here is to reconstruct novel views in between the two images. Since we would like our approach to have a low storage cost, we encode all the frames into a single neural network.\footnote{This is in contrast to using a separate network for each stereo frame.} As the number of frames $N$ can be large (90 for a 3 seconds video at 30 fps), we normalize the time coordinates to be between 0 and 1, i.e., $t = 0$ for the first frame and $t = 1$ for the last frame of the video. Moreover, we set the coordinates of the left and right views to $u_1 = -0.5$ and $u_2 = 0.5$, respectively.

Since our goal in this section is only view synthesis, our network needs to estimate a single channel Jacobian corresponding to the partial derivative of the horizontal displacement with respect to the view (i.e., disparity). With these \revadd{settings} we can optimize our network using Eq.~\ref{eq:xfields_loss} to perform this task. However, as shown in Fig.~\ref{fig:blendinset}, X-Fields is not able to produce satisfactory results. Next, we discuss our approaches to significantly improve the results.

\begin{figure}
\includegraphics[width=\linewidth]{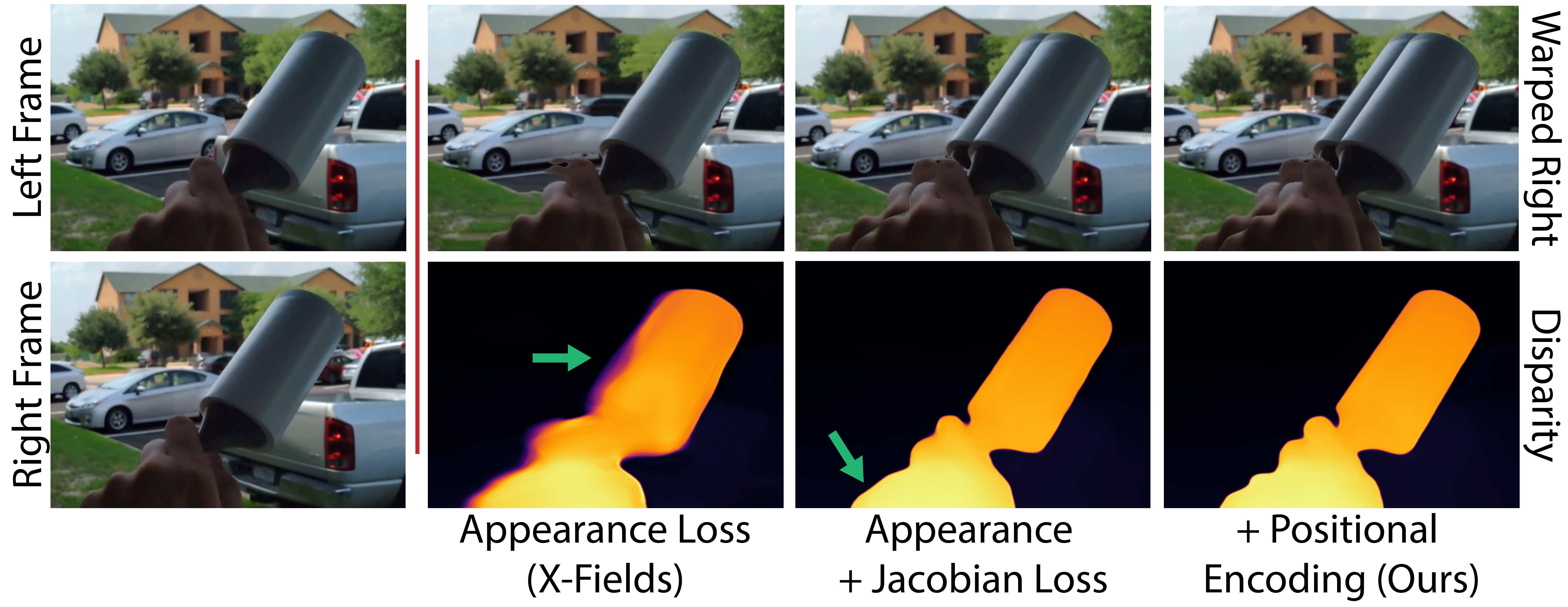}
\vspace{-0.25in}
\caption{We show the images captured from left and right views of a particular frame on the left. Using only the appearance loss, the network tries to reconstruct the occluded regions by stretching the flow outside the depth discontinuities. With our additional Jacobian loss, we supervise the Jacobian in the occluded regions and produce better maps. Through positional encoding, the network is able to properly learn the complex boundaries.}
\label{fig:warptrain}
\vspace{-0.25in}
\end{figure}

\vspace{-0.15in}
\paragraph{Jacobian Supervision:} We observe that with the extremely sparse inputs in our application (two views), the appearance loss in Eq.~\ref{eq:xfields_loss} does not provide sufficient supervision to estimate reliable Jacobians. Note that in this case, only one image can be used to reconstruct the image $f_\theta(\vect{y}_i)$ at coordinate $\vect{y}_i$ during optimization. For example, to reconstruct the left image at frame \revadd{$t=0$}%\revdel{0}
, $(u, t) = (-0.5, 0)$, we can only use the right image at that frame $(0.5, 0)$. Unfortunately, minimizing the appearance loss forces the network to reconstruct the occluded regions in one view from the other \revadd{(e.g., by texture stretching)} by estimating Jacobians that do not properly represent the disparity (see Fig.~\ref{fig:warptrain}). This negatively affect the quality of the interpolated results. 

To address this issue, we use an existing disparity estimation network (Li et al.~\cite{li2022practical} in our implementation) to constrain the estimated Jacobians in the occluded areas. Through training on a large number of scenes, this network is able to learn a prior and properly estimates the disparity in the occluded areas. Our key idea is to supervise our implicit network $g_\theta$ by only relying on the \emph{guiding} disparity (from the pre-trained network) in the occluded areas, but use both the appearance and disparity supervisions in the other areas. To do this, we introduce the following objective:

\vspace{-0.25in}
\begin{align}
    \label{eq:our_view_loss}
    \theta^* = \arg \min_\theta \sum_{j = 1}^2\sum_{i = 1}^N \Vert M^{\text{occ}} \odot \left(f_\theta(u_j, t_i) - f(u_j, t_i)\right) \Vert_1 \nonumber \\
    + \lambda \Vert g_\theta(u_j, t_i) - \tilde{J}(u_j, t_i)\Vert_1.
\end{align}
\vspace{-0.2in}

Here, $\tilde{J}(u_j, t_i)$ is the guiding Jacobian (disparity) and is obtained by passing the two views as the input to the pre-trained network. Specifically, to get $\tilde{J}(u_1, t_i)$ we pass the left and right images at frame $t_i$ as the input and reverse the order of the images to obtain the other Jacobian $\tilde{J}(u_2, t_i)$. Moreover, $M^{\text{occ}}$ is a binary mask with zero in the occluded areas and one in the other regions. We calculate this mask through forward-backward consistency check using the guiding Jacobians at the two views. This mask ensures that the warping loss (first term) is not used in the occluded areas. Finally, $\lambda$ defines the weight of the second term, which is set to $20/w$ (where $w$ is the frame width) in our implementation. In addition to improving the estimated Jacobians in the occluded areas (see Fig.~\ref{fig:warptrain}), the disparity supervision speeds up the convergence of the optimization and helps with the challenging cases \revadd{like} thin objects.

While the results produced with our method using the additional disparity supervision are reasonable, they sometimes contain ghosting and other artifacts around the depth discontinuities. We observe that these artifacts arise because the encoded Jacobians often lack sufficient details to represent the object boundaries (see Fig.~\ref{fig:warptrain}). This is mainly because of the difficulty in encoding a large number of frames into a single small network. To address this issue, we apply positional encoding~\cite{Mildenhall2020nerf} (with 5 frequencies) to the time coordinates and use them instead as the input to our implicit network. Note that while positional encoding is proposed for multi-layer perceptron (MLP) networks, we observe that it improves the quality of estimated Jacobians using coordinate-based convolutional network as well, as shown in Fig.~\ref{fig:warptrain}.

\begin{figure}
\includegraphics[width=\linewidth]{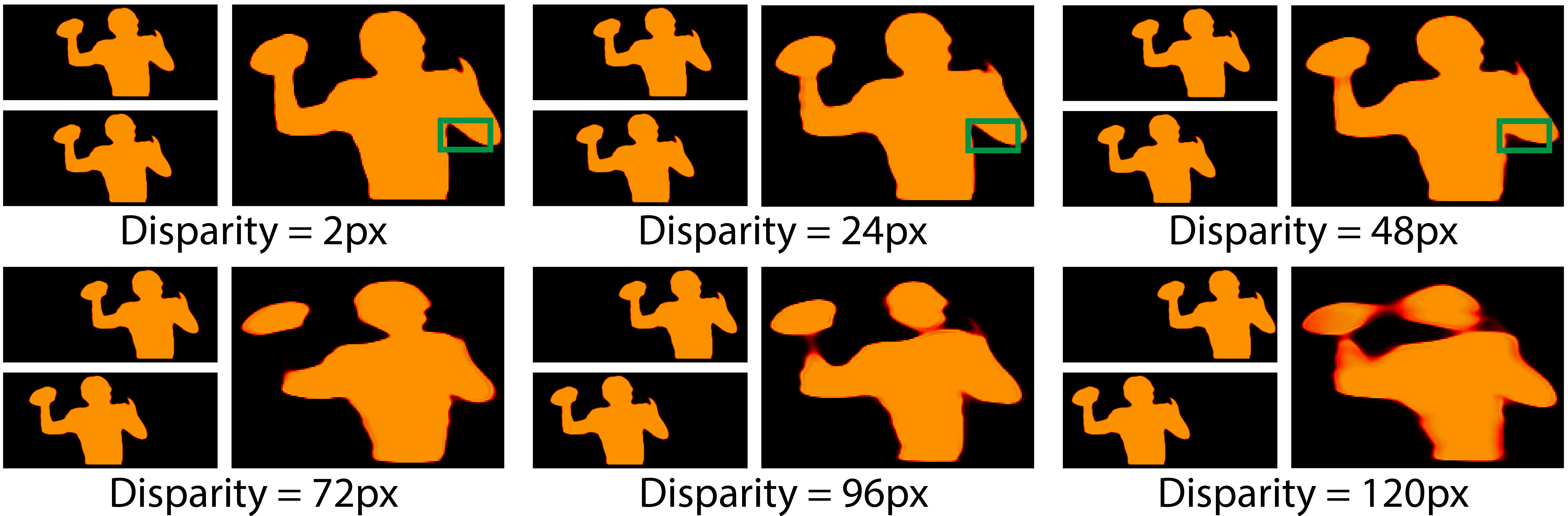}
\vspace{-0.3in}
\caption{We show the network's interpolation ability on examples with varying disparities. For each case, we show the left and right Jacobians on the left and the interpolated Jacobian (for the middle view) on the right. As seen the quality of the interpolated Jacobians starts to degrade for cases with 48 pixels disparity and above. Note that the gap under the arm (shown in green box) becomes smaller in the case with 48 pixels disparity.}
\label{fig:disp_shift_ab}
\vspace{-0.2in}
\end{figure}

\begin{figure}
\includegraphics[width=\linewidth]{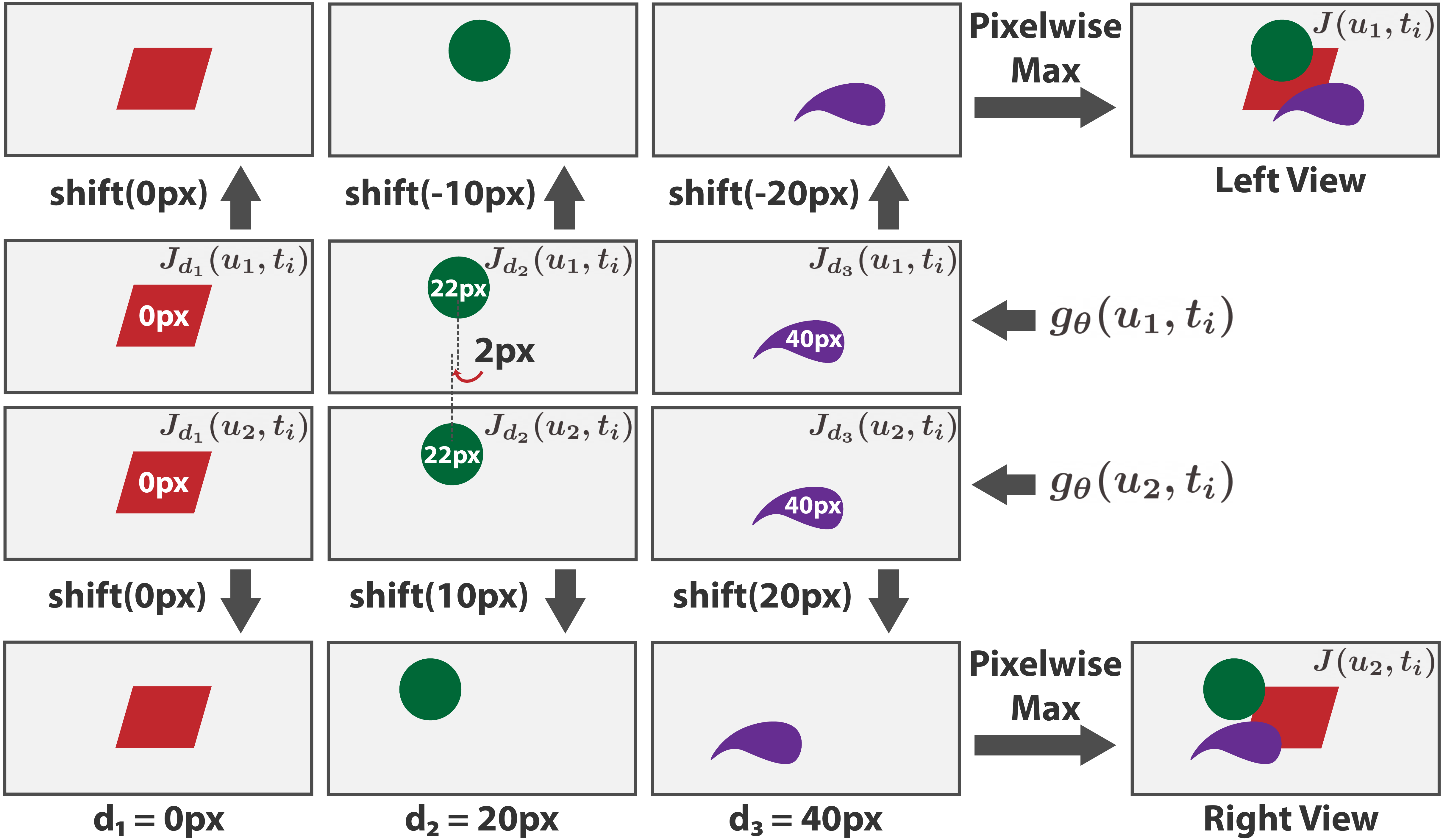}
\vspace{-0.3in}
\caption{We demonstrate our multi-plane disparities with three planes at 0, 20, and 40 pixels. Our network estimates Jacobians on three planes, each encoding the objects with disparities in the proximity of their pre-defined disparity. For example, the object with 22 pixels disparity (green circle) is encoded into the plane $d_2 = 20$. We reconstruct the final disparity by shifting each plane $d_k u$ pixels to the left and computing the pixel wise max on the shifted planes. For example, to reconstruct the Jacobian at the left view ($u_1 = -0.5$), we shift the plane at $d_2 = 20$ equal to -10 pixels to the left. Because we move the objects to their correct location using the shift, the objects in the encoded multi-plane Jacobians at the left and right views are spatially close. This significantly improves the interpolation ability of our network.}
\label{fig:multiplane_shift_viz}
\vspace{-0.3in}
\end{figure}

\vspace{-0.15in}
\paragraph{Multi-Plane Disparities:} This approach is able to produce high-quality Jacobians at intermediate views for cases where the maximum disparity in the scene is small (e.g., small baseline cameras). However, the quality of the interpolated Jacobians for cases with large disparities (e.g., large baseline cameras) deteriorates, as shown in Fig.~\ref{fig:disp_shift_ab}. This is because the objects with large disparities have a large spatial distance in the left and right Jacobians, making the implicit interpolation significantly more challenging. 

We address this issue by reducing the spatial distance between the objects in the encoded left and right Jacobians. Specifically, we propose multi-plane disparities to represent the Jacobians at each coordinate. As shown in Fig.~\ref{fig:multiplane_shift_viz}, instead of directly estimating the Jacobian, our network $g_\theta(u, t)$ estimates a set of Jacobians $J_{d_k}(u, v)$, each at a predefined disparity, $d_1, \cdots d_K$. The final Jacobian is then estimated from the multi-plane disparities as follows:

\vspace{-0.2in}
\begin{equation}
\label{eq:Jacob_Recon}
 J(u, t) = r(g_\theta(u, t)) = \max_{d_1, \cdots d_K} s(J_{d_k}(u, t), d_k u),
\end{equation}
\vspace{-0.2in}

\noindent where $s(J_{d_k}(u, t), d_k u)$ shifts the Jacobian at plane $k$, $J_{d_k}(u, t)$, $d_k u$ pixels to the left.

Since we use $u_1 = -0.5$ and $u_2 = 0.5$ as the left and right view coordinates, through this process, the left and right Jacobians at each plane, $J_{d_k}(u_1, t)$ and $J_{d_k}(u_2, t)$, are shifted equal to $d_k/2$ pixels in the opposite directions. Therefore, as shown in Fig.~\ref{fig:multiplane_shift_viz}, an object with $d_k$ pixels disparity will be at the same spatial location in the encoded left and right Jacobians at plane $k$, since the object will be moved to the correct location using the shift operator. This significantly enhances the interpolation quality as the network, in this case, encodes left and right multi-plane disparities that contain objects at small spatial distances.

Moreover, since the objects that are closer to the camera have larger disparities, we obtain the final Jacobian by selecting the \revadd{plane} with maximum disparity at each pixel (see Eq.~\ref{eq:Jacob_Recon}). % \revdel{By doing so, we only use a single plane to reconstruct the final Jacobian at each pixel.}
This allows the network to reconstruct the unselected regions in a desired manner. The additional flexibility, provided by the max operator, makes it superior to other choices, such as summation (see comparisons in the supplementary materials).

\begin{figure}
\includegraphics[width=\linewidth]{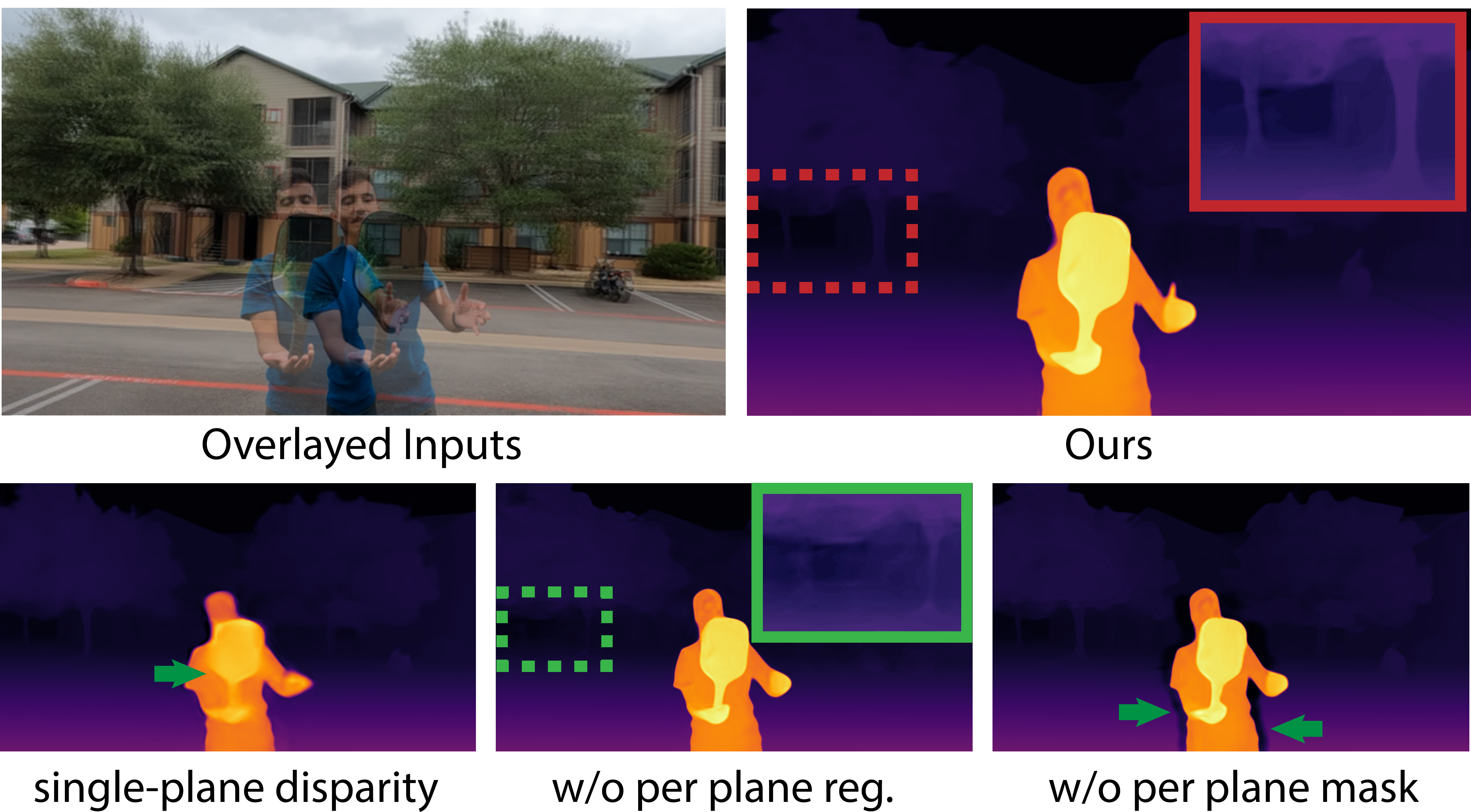}
\vspace{-0.3in}
\caption{We show the impact of our multi-plane disparity, as well as the per plane regularization term and \revadd{per plane} mask in Eq.~\ref{eq:our_view_loss_final}. \revadd{Single-plane disparity fails to reconstruct the details during interpolation. Without the per plane regularization, the network does not utilize all the planes effectively. Therefore, it is not able to produce reasonable results in the regions corresponding to the unused planes (background in this case). Not using the per plane mask leads to halo artifacts around foreground objects. See supplementary Fig.~\ref{fig:multiplane_supp} for more detailed intermediate results.}}
\label{fig:multiplane_loss_ab}
\vspace{-0.2in}
\end{figure}

Our network estimating the multi-plane disparities can be optimized using Eq.~\ref{eq:our_view_loss}, with a small modification; instead of directly estimating the Jacobian using the network, we use the network to estimate the multi-plane disparities and reconstruct the Jacobian using Eq.~\ref{eq:Jacob_Recon}. However, with such an optimization, there is no mechanism to enforce the network to utilize all the planes appropriately. For example, the network could potentially only use one of the planes to estimate the left and right Jacobians. In this case, our network will still not be able to properly interpolate the Jacobians, as shown in Fig.~\ref{fig:multiplane_loss_ab}.

\begin{figure}
\includegraphics[width=\linewidth]{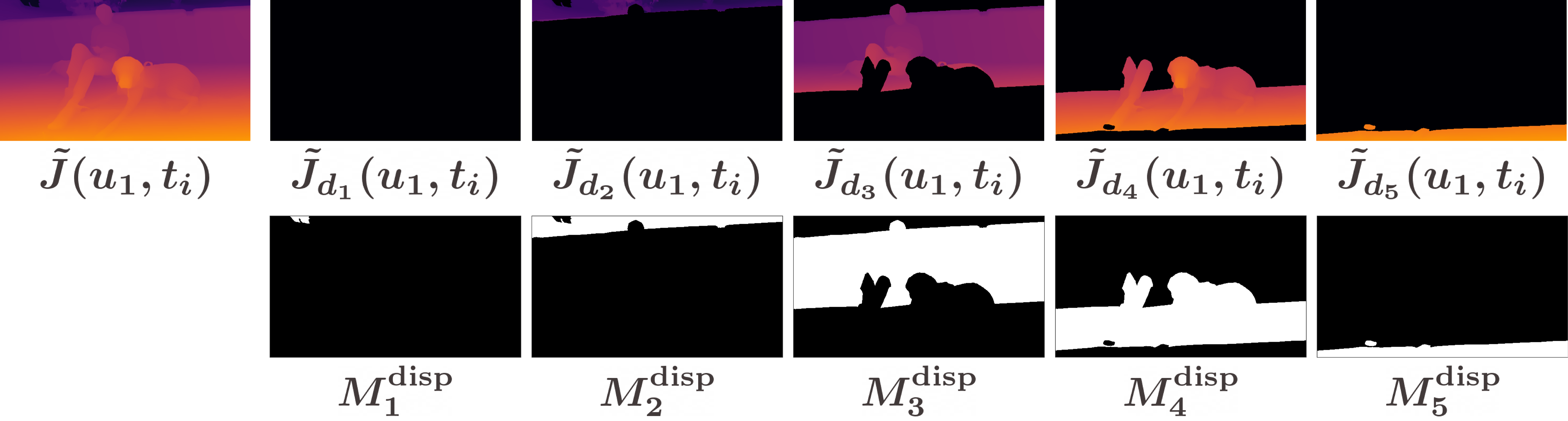}
\vspace{-0.3in}
\caption{On the top, we show the guiding disparity (leftmost) along with the per plane guiding disparity images. On the bottom, we show the corresponding masks for each plane.}
\label{fig:multiplane}
\vspace{-0.1in}
\end{figure}

To address this issue, we propose to apply a regularization loss between the estimated and guiding Jacobians at each plane. As shown in Fig.~\ref{fig:multiplane}, we compute the per plane guiding Jacobians by selecting disparities in the proximity of the plane's disparity. Specifically, to compute the guiding Jacobian at plane $k$ (i.e., $\tilde{J}_{d_k}(u_j, t_i)$), we only select the disparities in the range $d_k - l/2$ and $d_k + l/2$, where $l$ is the distance between consecutive disparities (e.g., 20 when the planes are at 0, 20, 40, etc.). Our final loss using this additional regularization is as follows:

\vspace{-0.20in}
\begin{align}
    \label{eq:our_view_loss_final}
    \theta^* = \arg \min_\theta \sum_{j = 1}^2\sum_{i = 1}^N \Vert M^{\text{occ}} \odot \left(f_\theta(u_j, t_i) - f(u_j, t_i)\right) \Vert_1 \nonumber \\
    + \lambda \Vert r(g_\theta(u_j, t_i)) - \tilde{J}(u_j, t_i)\Vert_1 \nonumber \\
    + \gamma \sum_{k = 1}^K\Vert M^{\text{disp}}_k \odot \left(s(J_{d_k}(u_j, t_i), d_k u_j) - \tilde{J}_{d_k}(u_j, t_i)\right)\Vert_1.
\end{align}
\vspace{-0.2in}

\noindent where $\gamma$ is the weight of the per plane regularization term and we set it to $1/w$ (where $w$ is the frame width) in our implementation. Note that we use the shifted estimated Jacobians in the regularization term to align them with the per plane guiding Jacobian. Moreover, we use a per plane mask $M^{\text{disp}}_k$ to ensure that this \revadd{regularization} is only applied in the regions where the guiding disparity has valid content. As shown in Fig.~\ref{fig:multiplane}, we obtain the binary per plane mask by setting the regions with valid content to one and the remaining areas to zero. Without this mask, the network will be forced to match the guiding Jacobians, even in the areas without a valid content. This negatively affects the quality of the results, as shown in Fig.~\ref{fig:multiplane_loss_ab}.

\begin{figure}
\includegraphics[width=\linewidth]{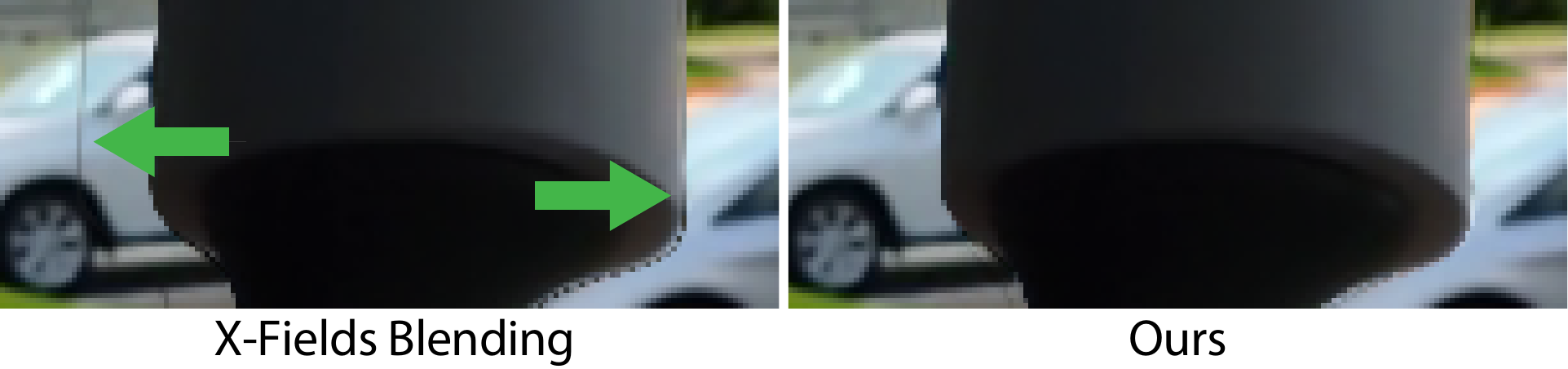}
\vspace{-0.3in}
\caption{We show the interpolated images generated using the X-Fields blending weights and our learned weights. Our approach does not have the distracting artifacts around the boundaries.}
\label{fig:occlusionmask}
\vspace{-0.25in}
\end{figure}

\vspace{-0.15in}
\paragraph{Blending:} Using the optimized network, we can obtain a Jacobian at any coordinate. This Jacobian can then be used to obtain the flows to the left and right views using Eq.~\ref{eq:flow}. The flows in turn can be used to warp the images to the novel coordinate. X-Fields~\cite{Bemana2020xfields} uses weight maps computed based on the forward-backward flow consistency to combine the images. However, their blending weights cannot fully avoid introducing the residual warping artifacts to the reconstructed image (see Fig.~\ref{fig:occlusionmask}). This is particularly a problem in stereo video interpolation since only one image is used to reconstruct the other view. As such the blending weights are not utilized during optimization.

To address this problem, we separately train a small network on a large dataset that takes the two inputs, warped images, as well as the corresponding flows and estimates a weight map. Using this map we obtain the reconstructed image at coordinate $(u, t_i)$ as:

\vspace{-0.1in}
\begin{equation}
\label{eq:blending}
    f_\theta(u, t_i) = \frac{(1 - c) W_l I_l + c W_r  I_r}{(1 - c) W_l + c W_r},
\end{equation}
\vspace{-0.1in}

\noindent where $c = u - u_0$, while $W_l$ is the estimated weight map for the left image and $W_r = 1 - W_l$. Moreover, $I_l$ and $I_r$ are the warped left and right views using the estimated Jacobian at coordinate $(u, t_i)$.

\subsection{Time Interpolation}
\label{ssec:TimeInterpolation}

The goal here is to reconstruct an image at a novel time coordinate from the two neighboring frames. Similar to the view synthesis case, we encode the entire stereo video into a single neural network to ensure low storage cost. Note that the network in this case estimates a two channel Jacobian at each coordinate, corresponding the partial derivative of the displacement in the $x$ and $y$ directions with respect to time. To handle this application, we apply all the enhancements from view synthesis (except multi-plane disparity, since it is specific to view synthesis), as we observe they improve the quality of the results. Specifically, we minimize the following loss, which is slightly different from the loss in Eq.~\ref{eq:our_view_loss}:

\vspace{-0.3in}
\begin{align}
    \label{eq:our_time_loss}
    \theta^* = \arg \min_\theta \sum_{j = 1}^2\sum_{i = 1}^N \Vert \left(f_\theta(u_j, t_i) - f(u_j, t_i)\right) \Vert_1 \nonumber \\
    + \lambda \Vert g_\theta(u_j, t_i) - \tilde{J}(u_j, t_i)\Vert_1.
\end{align}
\vspace{-0.2in}

The main differences here are that we do not have an occlusion mask $M^{\text{occ}}$ and our guidance Jacobian $\tilde{J}$ (estimated using Zhang et al.'s approach~\cite{Zhang2021separable}) has two channels. This flow supervision significantly enhances the results as the exposure in different frames varies slightly which makes it difficult for the appearance loss \revadd{(with brightness constancy assumption)} to find appropriate Jacobians. Note that we do not use the occlusion mask, since for time interpolation, the image at the novel coordinate is generated by combining the warped previous and next frames. Therefore, we assume that all the areas are visible at least in one of the neighboring frames. We also apply positional encoding to the time coordinates (10 frequencies) and use the learned blending weights as opposed to the weights computed by forward-backward flow consistency~\cite{Bemana2020xfields}.

\begin{figure}
\includegraphics[width=\linewidth]{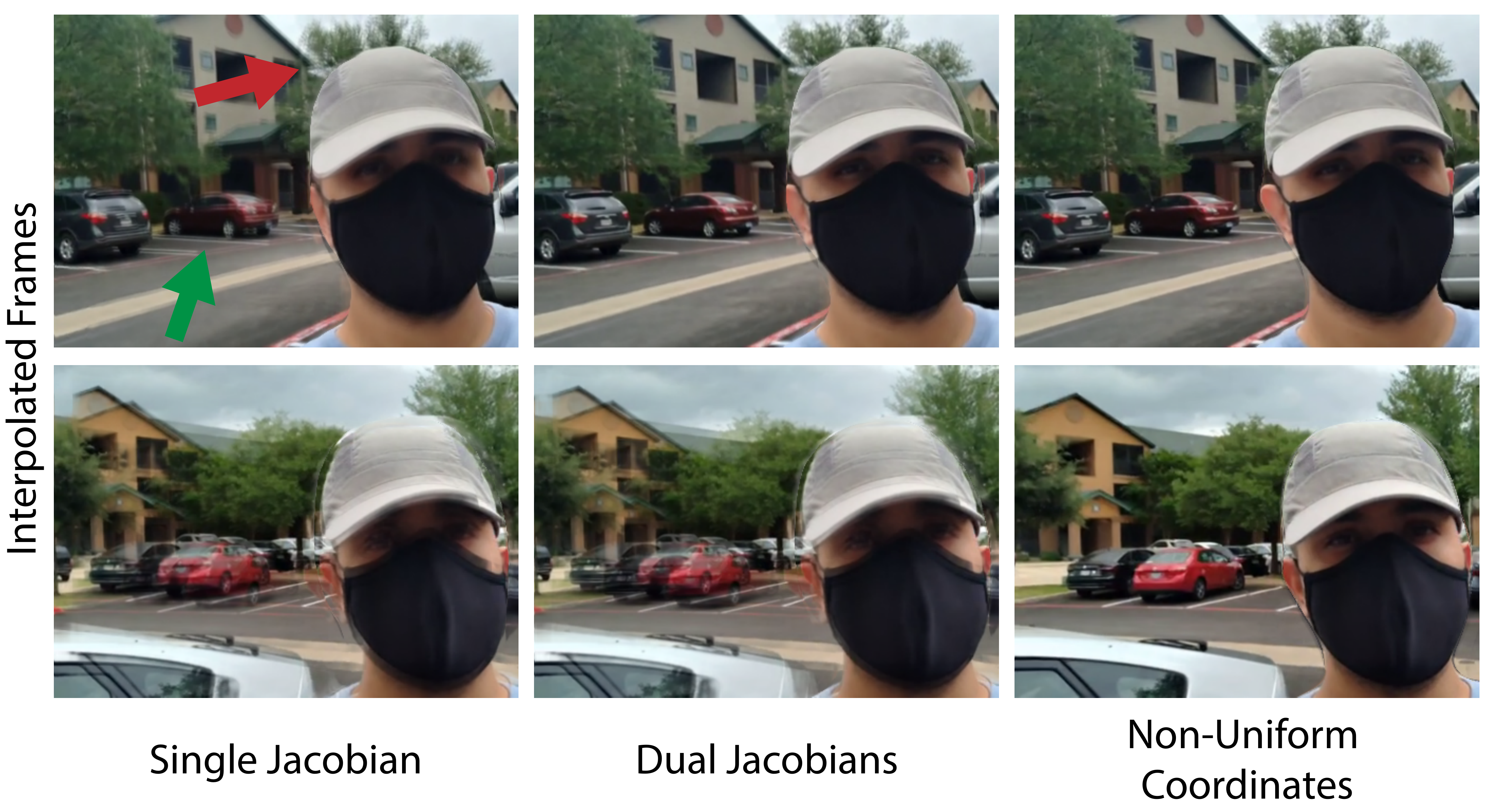}
\vspace{-0.3in}
\caption{We show two interpolated frames (top and bottom) generated using single Jacobian, dual jacobians, and our approach with non-uniform coordinates. On the top, we show that because of non-linearity of motion in the scene, with a single Jacobian we generate results with severe ghosting. This problem can be resolved with dual Jacobians. On the bottom, we show a frame with motion spikes. Both single and dual Jacobians fail to properly estimate the large motion in this case. Our approach with non-uniform coordinates, however, produces high-quality results in both cases.}
\label{fig:timeblend}
\vspace{-0.25in}
\end{figure}

Unfortunately, as shown in Fig.~\ref{fig:timeblend} (``Single Jacobian''), even with the additional enhancements, our system is not be able to produce satisfactory results in some cases. Specifically, we observe that the system struggles in cases where the motion becomes non-linear. This is because we use 
three consecutive frames during optimization; specifically, we minimize the error between $f(u_j, t_{i})$ and the reconstructed frame using the previous $f(u_j, t_{i-1})$ and next $f(u_j, t_{i+1})$ frames. The main assumption here is that the motion is linear, i.e., a single two channel Jacobian at $u_j, t_i$ can describe the flow to the previous and next frames. However, this assumption is typically violated in natural videos. 

\begin{figure}
\includegraphics[width=\linewidth]{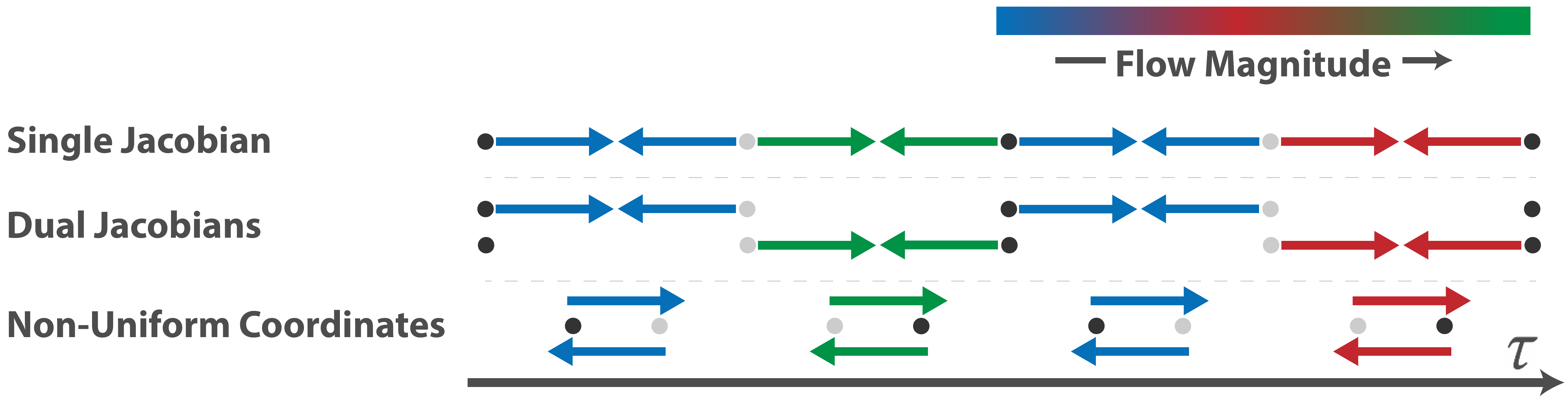}
\vspace{-0.25in}
\caption{We show the frames with a black or a gray dot and the two flows to the next and previous frames with two color coded arrows (color representing their magnitude). Note that we are illustrating the ideal flows that need to be encoded into the neural network. Natural videos have non-linear motions, and thus it is difficult to represent the two flows at each frame using a single Jacobian. A straightforward way to address this problem is to estimate two different Jacobians at each frame (dual Jacobians). However, as discussed in Sec.~\ref{ssec:TimeInterpolation}, this approach (the same as single Jacobian) will have difficulty handling motion spikes \revadd{(green)}. With our proposed non-uniform coordinates, we use two different coordinates to estimate the previous and next Jacobians at each frame. \revadd{The large unused regions in-between allow the network to smoothly accommodate motion spikes.}}
\label{fig:nonlinear}
\vspace{-0.53in}
\end{figure}

\begin{figure}
\includegraphics[width=\linewidth]{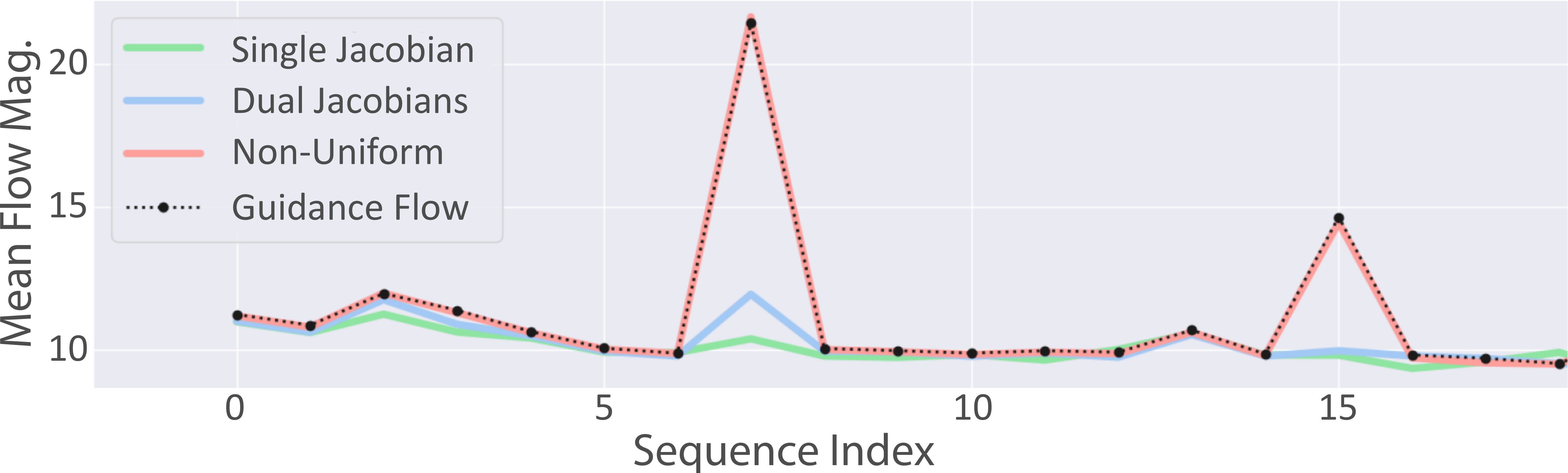}
\vspace{-0.3in}
\caption{We show the mean flow magnitude (obtained using a pre-trained flow network) for the consecutive frames in a video. As seen, natural videos often contain motion spikes, specially for handheld cameras. Single (X-Fields) and dual Jacobians are unable to properly estimate these spikes. Our method with non-uniform coordinates produces results that closely match the guidance flow.}
\label{fig:nonlinearchart}
\vspace{-0.2in}
\end{figure}

A straightforward way to handle this problem is to estimate two sets of Jacobians at each coordinate (see Fig.~\ref{fig:nonlinear}), $J_b(u_j, t_i)$ and $J_a(u_j, t_i)$, where they denote the Jacobians to the previous and next frames , respectively. The two Jacobians can be estimated using a single network (with a four channel output) or two separate networks. Although this approach can handle the cases with non-linear motions reasonably well (see ``Dual Jacobians'' in Fig.~\ref{fig:timeblend} - top), in some cases it produces results with severe ghosting (Fig.~\ref{fig:timeblend} - bottom). To understand the reason, we show the average flow magnitude of the guidance flow (obtained with a pre-trained network) for the consecutive frames of one view in Fig.~\ref{fig:nonlinearchart}. As seen, natural videos, in particular the ones captured with handheld devices, often contain motion spikes. Unfortunately, both the single and dual Jacobians solutions have difficulty producing Jacobians that can properly estimate such spikes. This is because the network smoothly interpolates between the observations, and thus the spikes will be over-smoothed.

We address this issue by proposing to encode the Jacobians using non-uniform time coordinates $\tau$ (see Fig.~\ref{fig:nonlinear}). Specifically, we estimate all the Jacobians to the next frame ($J_a$) at the original time coordinate ($\tau = t$ for $J_a$). However, for the Jacobian to the previous frame ($J_b$) we shift the coordinate closer to the previous frame, i.e., $\tau = t - \alpha$ for $J_b$, where $\alpha$ is set to 0.9 in our implementation. With this transformation, we are able to effectively encode the Jacobians (see Fig.~\ref{fig:nonlinearchart}) and produce high-quality interpolated frames (Fig.~\ref{fig:timeblend}). In this case, \revadd{the large unused regions in-between allow the network to smoothly ramp up and down to and from the motion spikes.}

\subsection{View-Time Synthesis}
\label{ssec:view_time}

Our goal here is to combine the view and time interpolation systems to be able to synthesize an image at any novel view-time coordinates given a stereo video. While we can perform both view and time synthesis by encoding the view and time Jacobians into one network, we observe that the two types of Jacobians often conflict with each other. This, unfortunately, negatively affects the quality of the results.

\begin{figure}
\includegraphics[width=\linewidth]{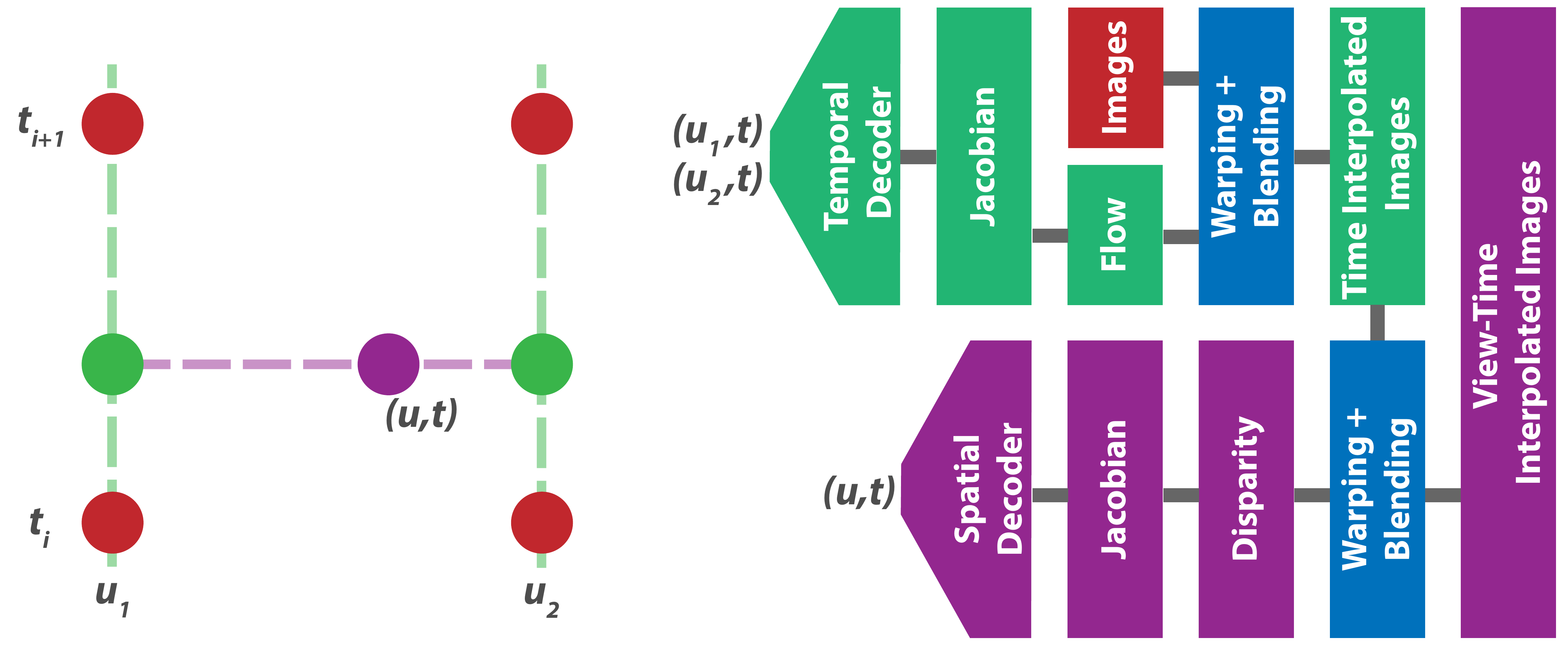}
\vspace{-0.25in}
\caption{Overview of our view-time interpolation system.}
\vspace{-0.25in}
\label{fig:viewtimeviz}
\end{figure}

Therefore, we propose to perform the view-time synthesis in two stages (see Fig.~\ref{fig:viewtimeviz}). Given 4 frames neighboring the coordinate of interest $(u, t)$, we first obtain the time interpolated images at coordinates $(u_1, t)$ and $(u_2, t)$. In the second stage, we perform view interpolation between these two images to calculate the final image at coordinate $(u, t)$.

\section{Implementation}

% \revdel{In this section, we discuss the necessary details to implement our approach.}

We set the coordinates of the left and right views to $u_1 = -0.5$ and $u_2 = 0.5$, respectively. However, when passing these view coordinates to the network we scale them down by a factor of 30 (i.e., $u_1 = \revadd{-1/60}$ and $u_2 = \revadd{1/60}$), as we \revadd{empirically} observe that with closer coordinates, the network is able to better interpolate between the Jacobians. This is because with closer coordinates the network tends to produce intermediate Jacobians that are correlated with the ones at the left and right views. \revadd{However, after a certain point, the interpolation quality deteriorates as we bring the coordinates closer~\cite{Figueiredo2023frame}}.

We use the convolutional decoder network as proposed by Bemana et al.~\cite{Bemana2020xfields} with a capacity factor of 16 for both view and time interpolations. Moreover, we use a UNet with 5 downsampling/upsampling layers to estimate the blending weights. %This network takes a 16 channel input and estimates a one channel weight map ($16\rightarrow16\rightarrow32\rightarrow64\rightarrow32\rightarrow16\rightarrow1$). We use a LeakyReLU activation function for all the layers except the output for which we use sigmoid. 
We train the network for 10k iterations using the Adam optimizer~\cite{Kingma15adam} with a learning rate of $10^{-4}$ on the Vimeo90K~\cite{Xue2019video} video dataset. As the loss, we use the L1 distance between the blended and ground truth images.

In our implementation, we further enhance the view weight maps ($W_l$ and $W_r$ in Eq.~\ref{eq:blending}) by computing gradient of the flow in the $x$ direction. We then threshold the derivative to identify the edge regions. We expand these edges horizontally by applying dilation-erosion (morphological operations) with an anisotropic kernel to create a residual mask. Finally, we combine this residual mask with the weight map through binary operations.

We perform our per-scene optimization for 100k iterations using Adam. We start with a learning rate of $10^{-4}$ and decrease it by a factor of 0.4 every 33k iterations to speed up convergence.

\section{Results}
\label{sec:Results}

\begin{figure*}
\includegraphics[width=\linewidth]{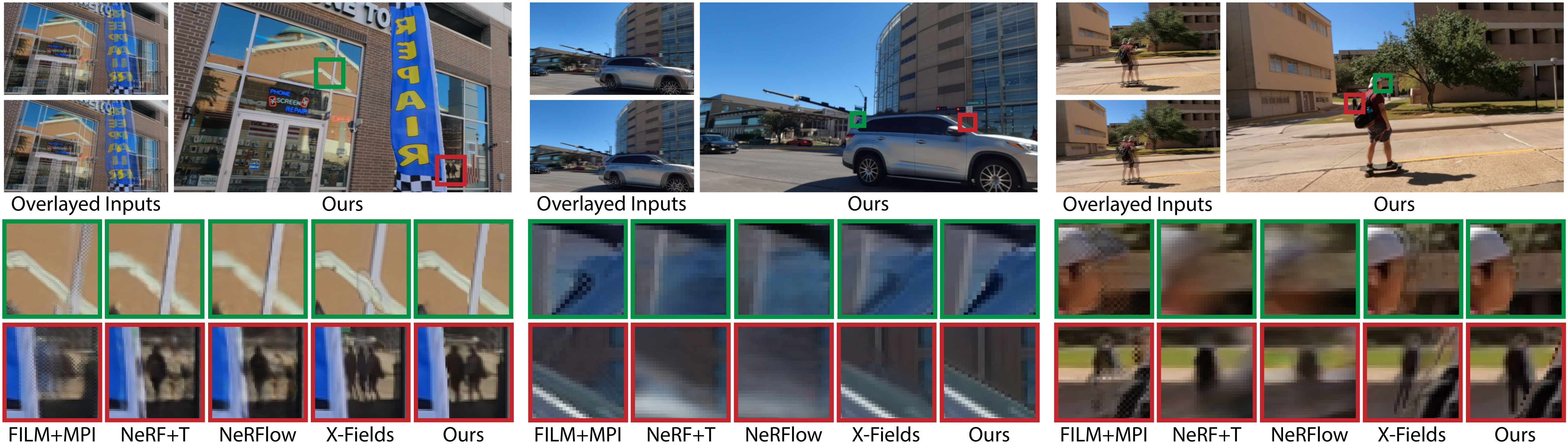}
\vspace{-0.2in}
\caption{Comparison against several state-of-the-art methods on view-time interpolation. On the left we show the overlayed left and right views for two consecutive frames neighboring the coordinate of interest.}
\label{fig:viewtime_results}
\vspace{-0.2in}
\end{figure*}

We compare our approach against the approaches by Bemana et al.~\cite{Bemana2020xfields} (X-Fields) and Du et al.~\cite{Du2021nerflow} (NeRF+T, NeRFlow). We also compare against a combination of MPI (for view interpolation) by  Zhou et al.~\cite{Zhou2018stereo} and FILM by Reda et al.~\cite{reda2022film} (for time interpolation). We call this combined method FILM+MPI throughout this section. For X-Fields, we use a larger network than ours to ensure the network capacity is not a limiting factor for their performance. Du et al. first train a NeRF with an additional input for the time coordinate (NeRF+T) and then fine-tune it for dynamic scenes using pre-trained flows and depth maps (NeRFlow). We compare our method against both versions of this approach. For all the approaches, we use the source code provided by the authors. Here, we show results on a few scenes, but more comparisons, and the videos, are provided in the supplementary materials.

\subsection{Qualitative Results}
We capture a set of stereo videos with a variety of motions using a stereo GoPro camera rig. We show the results using this camera setup in the paper, but we also test our approach using Lume Pad~\cite{leia} and provide the results in the supplementary video.

We show comparisons against several state-of-the art approaches on a few scenes in Fig.~\ref{fig:viewtime_results}. For all the scenes, we show view-time interpolation at the middle of four observed view-time frames. As seen, other approaches produce results with noticeable ghosting and other artifacts, while our results are sharp and have clear boundaries.

\subsection{Quantitative Results}
We numerically compare our approach against the other methods on two lightfield video datasets, Sintel~\cite{Kinoshita2021depth} and LFVID~\cite{Sabater2017dataset}, in terms of PSNR, SSIM~\cite{Wang2004image}, and LPIPS~\cite{Zhang2018the}. The comparisons against FILM+MPI and X-Fields are shown in Table~\ref{tab:ResultsViewTime}. As seen, our approach produces significantly better results than both of these methods across all the metrics. Note that, we excluded NeRF+T and NeRFlow from these comparisons as the camera calibration fails for some of the sequences (some scenes are captured with a tripod mounted camera).

\begin{table}[!t]
\renewcommand{\arraystretch}{1.3}
\caption{View-time synthesis results on Sintel~\cite{Kinoshita2021depth} and LFVID~\cite{Sabater2017dataset} datasets.}
\centering
\vspace{-0.1in}
\begin{footnotesize}
\begin{tabular}{l|ccc|ccc}
  \hline\hline
  & \multicolumn{3}{c|}{Sintel} & \multicolumn{3}{c}{LFVID}\\
  & PSNR & SSIM & LPIPS & PSNR & SSIM & LPIPS\\
  \hline
  FILM+MPI & 24.39 & 0.782 & 0.1423 & 18.23 & 0.496 & 0.3046\\
  X-Fields & 25.38 & 0.814 & 0.1263 & 19.33 & 0.552 & 0.2365\\
  % OursOld & \textbf{26.21} & \textbf{0.850} & \textbf{0.1038} & \textbf{21.21} & \textbf{0.610} & \textbf{0.1756}\\
  Ours & \textbf{26.33} & \textbf{0.860} & \textbf{0.0989} & \textbf{23.49} & \textbf{0.696} & \textbf{0.1345}\\
  \hline\hline
\end{tabular}
\vspace{-0.2in}
\label{tab:ResultsViewTime}
\end{footnotesize}
\end{table}

\subsection{Training and Inference Times}

We compare the training (optimization) and inference speed of our approach against the other methods. The training times are obtained on a machine with an NVIDIA V100 GPU, while the inference speed is measured on a machine with a 2080Ti GPU. We provide the timings for a stereo video with 100 360p frames. Our approach takes roughly 120 minutes to train our view and time interpolation networks. In comparison, X-Fields requires 170 minutes of training, while NeRF+T and NeRFlow take around 5 and 25 hours, respectively. At run-time, our approach takes 44 ms to generate a novel view-time image. In contrast, X-Fields takes 26 ms, while NeRF+T and NeRFlow require 17 seconds to generate a single image. Note that, X-Fields is faster as it performs view-time synthesis using a single network, while ours requires evaluating the Jacobians using two separate networks. However, we believe given the improvement in quality, this is justified. Nevertheless, the performance can be significantly improved by optimizing the network architecture and implementing the approach on efficient frameworks like \emph{tiny-cuda-nn}~\cite{tiny-cuda-nn}.

Note that, MPI does not perform any optimization and can render novel images in real-time given the MPI for each frame. However, the MPI for each frame is around 120 MB (12 GB for 100 frames), and thus their method is significantly storage and memory intensive. In comparison, our network takes around 12 MB of storage space for the entire 100 frames Additionally, MPI only performs view synthesis and we augmented this method with FILM, which takes roughly 0.5s per frame, to be able to handle view-time interpolation. While there are other approaches, such as RIFE~\cite{huang2022rife}, that can perform time interpolation at real-time, they often produce results with lower quality.

\subsection{Limitations}
Our approach uses Jacobian supervision, and thus the performance of our system depends on the quality of the guidance Jacobians (estimated with Li et al.~\cite{li2022practical} and Zhang et al.~\cite{Zhang2021separable}). As such, poor quality guidance Jacobians can negatively affect our optimization. However, as discussed, pure per-scene optimization for such an ill-posed problem is not effective and incorporating the results of networks, trained over a large number of scenes is necessary. Additionally, our method will not be able to properly reconstruct regions that are occluded in both neighboring frames, e.g., left and right views. Such information, however, might exist in other frames in the video. Therefore, addressing this problem by combining the NeRF-based approaches with our method would be an interesting future research.

\section{Conclusion}
\label{sec:Conclusion}
We present an approach to generate images from any novel view-time coordinates from an input stereo video. We analyze and identify the problems with using X-Fields in our application. We make two key observations based on our analysis: 1) the network struggles to interpolate the Jacobians for cases with large disparities and 2) the main assumption of X-Fields is linear motion which is violated in natural videos. Based on these observations, we propose multi-plane disparities and non-uniform time coordinates to improve the results. We demonstrate that our method significantly outperforms the state of the art.

\section{Acknowledgement}
\label{sec:Acknowledgement}
\revadd{The authors would like to thank the reviewers for their comments and suggestions. This work was funded by Leia Inc. (contract \#415290).}

%%%%%%%%% REFERENCES
{\small
\bibliographystyle{ieee_fullname}
\bibliography{main}

\begin{thebibliography}{10}\itemsep=-1pt

\bibitem{Bao2019depth}
Wenbo Bao, Wei-Sheng Lai, Chao Ma, Xiaoyun Zhang, Zhiyong Gao, and Ming-Hsuan
  Yang.
\newblock Depth-aware video frame interpolation.
\newblock In {\em Proceedings of the IEEE/CVF Conference on Computer Vision and
  Pattern Recognition (CVPR)}, June 2019.

\bibitem{Bemana2020xfields}
Mojtaba Bemana, Karol Myszkowski, Hans-Peter Seidel, and Tobias Ritschel.
\newblock X-fields: Implicit neural view-, light- and time-image interpolation.
\newblock {\em ACM Trans. Graph. (Proc. SIGGRAPH Asia)}, 39(6), 2020.

\bibitem{broxton2020immersive}
Michael Broxton, John Flynn, Ryan Overbeck, Daniel Erickson, Peter Hedman,
  Matthew DuVall, Jason Dourgarian, Jay Busch, Matt Whalen, and Paul Debevec.
\newblock Immersive light field video with a layered mesh representation.
\newblock {\em ACM Transactions on Graphics (Proc. SIGGRAPH)},
  39(4):86:1--86:15, 2020.

\bibitem{Beuhler2001unstructured}
Chris Buehler, Michael Bosse, Leonard McMillan, Steven Gortler, and Michael
  Cohen.
\newblock Unstructured lumigraph rendering.
\newblock In {\em Proceedings of the 28th Annual Conference on Computer
  Graphics and Interactive Techniques}, SIGGRAPH '01, page 425–432, 2001.

\bibitem{Chaurasia2013depth}
Gaurav Chaurasia, Sylvain Duchene, Olga Sorkine-Hornung, and George Drettakis.
\newblock Depth synthesis and local warps for plausible image-based navigation.
\newblock {\em ACM Trans. Graph.}, 32(3), 2013.

\bibitem{Choi2020scene}
Myungsub Choi, Janghoon Choi, Sungyong Baik, Tae~Hyun Kim, and Kyoung~Mu Lee.
\newblock Scene-adaptive video frame interpolation via meta-learning.
\newblock In {\em Proceedings of the IEEE/CVF Conference on Computer Vision and
  Pattern Recognition (CVPR)}, June 2020.

\bibitem{choi2020channel}
Myungsub Choi, Heewon Kim, Bohyung Han, Ning Xu, and Kyoung~Mu Lee.
\newblock Channel attention is all you need for video frame interpolation.
\newblock In {\em AAAI}, 2020.

\bibitem{Debevec1996modeling}
Paul~E. Debevec, Camillo~J. Taylor, and Jitendra Malik.
\newblock Modeling and rendering architecture from photographs: A hybrid
  geometry- and image-based approach.
\newblock In {\em Proceedings of the 23rd Annual Conference on Computer
  Graphics and Interactive Techniques}, SIGGRAPH '96, page 11–20, 1996.

\bibitem{kangle2021depth}
Kangle Deng, Andrew Liu, Jun-Yan Zhu, and Deva Ramanan.
\newblock Depth-supervised {NeRF}: Fewer views and faster training for free.
\newblock In {\em Proceedings of the IEEE/CVF Conference on Computer Vision and
  Pattern Recognition (CVPR)}, June 2022.

\bibitem{Du2021nerflow}
Yilun Du, Yinan Zhang, Hong-Xing Yu, Joshua~B. Tenenbaum, and Jiajun Wu.
\newblock Neural radiance flow for 4d view synthesis and video processing.
\newblock In {\em Proceedings of the IEEE/CVF International Conference on
  Computer Vision}, 2021.

\bibitem{Figueiredo2023frame}
Pedro Figueir\^edo, Avinash Paliwal, and Nima~Khademi Kalantari.
\newblock Frame interpolation for dynamic scenes with implicit flow encoding.
\newblock In {\em Proceedings of the IEEE/CVF Winter Conference on Applications
  of Computer Vision (WACV)}, pages 218--228, January 2023.

\bibitem{Flynn2019deepview}
John Flynn, Michael Broxton, Paul Debevec, Matthew DuVall, Graham Fyffe, Ryan
  Overbeck, Noah Snavely, and Richard Tucker.
\newblock Deepview: View synthesis with learned gradient descent.
\newblock In {\em Proceedings of the IEEE/CVF Conference on Computer Vision and
  Pattern Recognition (CVPR)}, June 2019.

\bibitem{Flynn_2016}
John Flynn, Ivan Neulander, James Philbin, and Noah Snavely.
\newblock Deep stereo: Learning to predict new views from the world's imagery.
\newblock In {\em 2016 IEEE Conference on Computer Vision and Pattern
  Recognition (CVPR)}, pages 5515--5524, 2016.

\bibitem{Gao2021Dynamic}
Chen Gao, Ayush Saraf, Johannes Kopf, and Jia-Bin Huang.
\newblock Dynamic view synthesis from dynamic monocular video.
\newblock In {\em Proceedings of the IEEE International Conference on Computer
  Vision}, 2021.

\bibitem{Hedman2017casual}
Peter Hedman, Suhib Alsisan, Richard Szeliski, and Johannes Kopf.
\newblock Casual 3d photography.
\newblock {\em ACM Trans. Graph.}, 36(6), 2017.

\bibitem{Hedman2018instant}
Peter Hedman and Johannes Kopf.
\newblock Instant 3d photography.
\newblock {\em ACM Trans. Graph.}, 37(4), 2018.

\bibitem{Hedman2021baking}
Peter Hedman, Pratul~P. Srinivasan, Ben Mildenhall, Jonathan~T. Barron, and
  Paul Debevec.
\newblock Baking neural radiance fields for real-time view synthesis.
\newblock In {\em Proceedings of the IEEE/CVF International Conference on
  Computer Vision (ICCV)}, pages 5875--5884, October 2021.

\bibitem{huang2022rife}
Zhewei Huang, Tianyuan Zhang, Wen Heng, Boxin Shi, and Shuchang Zhou.
\newblock Real-time intermediate flow estimation for video frame interpolation.
\newblock In {\em Proceedings of the European Conference on Computer Vision
  (ECCV)}, 2022.

\bibitem{Jain2021putting}
Ajay Jain, Matthew Tancik, and Pieter Abbeel.
\newblock Putting nerf on a diet: Semantically consistent few-shot view
  synthesis.
\newblock In {\em Proceedings of the IEEE/CVF International Conference on
  Computer Vision (ICCV)}, pages 5885--5894, October 2021.

\bibitem{Jiang2018super}
Huaizu Jiang, Deqing Sun, Varun Jampani, Ming-Hsuan Yang, Erik Learned-Miller,
  and Jan Kautz.
\newblock Super slomo: High quality estimation of multiple intermediate frames
  for video interpolation.
\newblock In {\em Proceedings of the IEEE Conference on Computer Vision and
  Pattern Recognition (CVPR)}, June 2018.

\bibitem{Kalantari_2016}
Nima~Khademi Kalantari, Ting-Chun Wang, and Ravi Ramamoorthi.
\newblock Learning-based view synthesis for light field cameras.
\newblock {\em ACM Transactions on Graphics (Proceedings of SIGGRAPH Asia
  2016)}, 35(6), 2016.

\bibitem{kalluri2021flavr}
Tarun Kalluri, Deepak Pathak, Manmohan Chandraker, and Du Tran.
\newblock Flavr: Flow-agnostic video representations for fast frame
  interpolation.
\newblock {\em arXiv preprint arXiv:2012.08512}, 2020.

\bibitem{Kingma15adam}
Diederik Kingma and Jimmy Ba.
\newblock Adam: A method for stochastic optimization.
\newblock In {\em International Conference on Learning Representations (ICLR)},
  2015.

\bibitem{Kinoshita2021depth}
Takahiro Kinoshita and Satoshi Ono.
\newblock Depth estimation from 4d light field videos.
\newblock In {\em International Workshop on Advanced Imaging Technology (IWAIT)
  2021}, volume 11766. International Society for Optics and Photonics, 2021.

\bibitem{Klose2015sampling}
Felix Klose, Oliver Wang, Jean-Charles Bazin, Marcus Magnor, and Alexander
  Sorkine-Hornung.
\newblock Sampling based scene-space video processing.
\newblock {\em ACM Trans. Graph.}, 34(4), 2015.

\bibitem{leia}
Leia.
\newblock Leia inc.
\newblock \url{https://www.leiainc.com/}, 2022.

\bibitem{li2022practical}
Jiankun Li, Peisen Wang, Pengfei Xiong, Tao Cai, Ziwei Yan, Lei Yang, Jiangyu
  Liu, Haoqiang Fan, and Shuaicheng Liu.
\newblock Practical stereo matching via cascaded recurrent network with
  adaptive correlation.
\newblock In {\em Proceedings of the IEEE/CVF Conference on Computer Vision and
  Pattern Recognition}, pages 16263--16272, 2022.

\bibitem{Li2020synthesizing}
Qinbo Li and Nima Khademi~Kalantari.
\newblock Synthesizing light field from a single image with variable mpi and
  two network fusion.
\newblock {\em ACM Transactions on Graphics}, 39(6), 12 2020.

\bibitem{Li2020neural}
Zhengqi Li, Simon Niklaus, Noah Snavely, and Oliver Wang.
\newblock Neural scene flow fields for space-time view synthesis of dynamic
  scenes.
\newblock In {\em Proceedings of the IEEE/CVF Conference on Computer Vision and
  Pattern Recognition (CVPR)}, 2021.

\bibitem{lin2021deep}
Kai-En Lin, Lei Xiao, Feng Liu, Guowei Yang, and Ravi Ramamoorthi.
\newblock Deep 3d mask volume for view synthesis of dynamic scenes.
\newblock In {\em ICCV}, 2021.

\bibitem{mildenhall2019local}
Ben Mildenhall, Pratul~P. Srinivasan, Rodrigo Ortiz-Cayon, Nima~Khademi
  Kalantari, Ravi Ramamoorthi, Ren Ng, and Abhishek Kar.
\newblock Local light field fusion: Practical view synthesis with prescriptive
  sampling guidelines.
\newblock {\em ACM Transactions on Graphics (TOG)}, 2019.

\bibitem{Mildenhall2020nerf}
Ben Mildenhall, Pratul~P. Srinivasan, Matthew Tancik, Jonathan~T. Barron, Ravi
  Ramamoorthi, and Ren Ng.
\newblock Nerf: Representing scenes as neural radiance fields for view
  synthesis.
\newblock In {\em ECCV}, 2020.

\bibitem{tiny-cuda-nn}
Thomas M\"uller.
\newblock Tiny {CUDA} neural network framework, 2021.
\newblock https://github.com/nvlabs/tiny-cuda-nn.

\bibitem{Niklaus2018context}
Simon Niklaus and Feng Liu.
\newblock Context-aware synthesis for video frame interpolation.
\newblock In {\em Proceedings of the IEEE conference on computer vision and
  pattern recognition}, pages 1701--1710, 2018.

\bibitem{Niklaus2020softmax}
Simon Niklaus and Feng Liu.
\newblock Softmax splatting for video frame interpolation.
\newblock In {\em Proceedings of the IEEE/CVF Conference on Computer Vision and
  Pattern Recognition (CVPR)}, June 2020.

\bibitem{park2020bmbc}
Junheum Park, Keunsoo Ko, Chul Lee, and Chang-Su Kim.
\newblock Bmbc: Bilateral motion estimation with bilateral cost volume for
  video interpolation.
\newblock In {\em European Conference on Computer Vision}, pages 109--125,
  2020.

\bibitem{park2021nerfies}
Keunhong Park, Utkarsh Sinha, Jonathan~T. Barron, Sofien Bouaziz, Dan~B
  Goldman, Steven~M. Seitz, and Ricardo Martin-Brualla.
\newblock Nerfies: Deformable neural radiance fields.
\newblock {\em ICCV}, 2021.

\bibitem{park2021hypernerf}
Keunhong Park, Utkarsh Sinha, Peter Hedman, Jonathan~T. Barron, Sofien Bouaziz,
  Dan~B Goldman, Ricardo Martin-Brualla, and Steven~M. Seitz.
\newblock Hypernerf: A higher-dimensional representation for topologically
  varying neural radiance fields.
\newblock {\em ACM Trans. Graph.}, 40(6), dec 2021.

\bibitem{Peng2021neural}
Sida Peng, Yuanqing Zhang, Yinghao Xu, Qianqian Wang, Qing Shuai, Hujun Bao,
  and Xiaowei Zhou.
\newblock Neural body: Implicit neural representations with structured latent
  codes for novel view synthesis of dynamic humans.
\newblock In {\em Proceedings of the IEEE/CVF Conference on Computer Vision and
  Pattern Recognition (CVPR)}, pages 9054--9063, June 2021.

\bibitem{Penner2017soft}
Eric Penner and Li Zhang.
\newblock Soft 3d reconstruction for view synthesis.
\newblock {\em ACM Trans. Graph.}, 36(6), 2017.

\bibitem{Pumarola2021dnerf}
Albert Pumarola, Enric Corona, Gerard Pons-Moll, and Francesc Moreno-Noguer.
\newblock D-nerf: Neural radiance fields for dynamic scenes.
\newblock In {\em Proceedings of the IEEE/CVF Conference on Computer Vision and
  Pattern Recognition (CVPR)}, pages 10318--10327, June 2021.

\bibitem{reda2022film}
Fitsum Reda, Janne Kontkanen, Eric Tabellion, Deqing Sun, Caroline Pantofaru,
  and Brian Curless.
\newblock Film: Frame interpolation for large motion.
\newblock In {\em European Conference on Computer Vision (ECCV)}, 2022.

\bibitem{Reda2019unsupervised}
Fitsum~A Reda, Deqing Sun, Aysegul Dundar, Mohammad Shoeybi, Guilin Liu,
  Kevin~J Shih, Andrew Tao, Jan Kautz, and Bryan Catanzaro.
\newblock Unsupervised video interpolation using cycle consistency.
\newblock In {\em The IEEE International Conference on Computer Vision (ICCV)},
  October 2019.

\bibitem{Sabater2017dataset}
Neus Sabater, Guillaume Boisson, Benoit Vandame, Paul Kerbiriou, Frederic
  Babon, Matthieu Hog, Tristan Langlois, Remy Gendrot, Olivier Bureller, Arno
  Schubert, and Valerie Allie.
\newblock Dataset and pipeline for multi-view light-field video.
\newblock In {\em CVPR Workshops}, 2017.

\bibitem{Snavely2006photo}
Noah Snavely, Steven~M. Seitz, and Richard Szeliski.
\newblock Photo tourism: Exploring photo collections in 3d.
\newblock {\em ACM Trans. Graph.}, 25(3):835–846, 2006.

\bibitem{Srinivasan2021nerv}
Pratul~P. Srinivasan, Boyang Deng, Xiuming Zhang, Matthew Tancik, Ben
  Mildenhall, and Jonathan~T. Barron.
\newblock Nerv: Neural reflectance and visibility fields for relighting and
  view synthesis.
\newblock In {\em Proceedings of the IEEE/CVF Conference on Computer Vision and
  Pattern Recognition (CVPR)}, pages 7495--7504, June 2021.

\bibitem{Srinivasan2019pushing}
Pratul~P. Srinivasan, Richard Tucker, Jonathan~T. Barron, Ravi Ramamoorthi, Ren
  Ng, and Noah Snavely.
\newblock Pushing the boundaries of view extrapolation with multiplane images.
\newblock In {\em Proceedings of the IEEE/CVF Conference on Computer Vision and
  Pattern Recognition (CVPR)}, June 2019.

\bibitem{Tretschk2021nonrigid}
Edgar Tretschk, Ayush Tewari, Vladislav Golyanik, Michael Zollh\"ofer,
  Christoph Lassner, and Christian Theobalt.
\newblock Non-rigid neural radiance fields: Reconstruction and novel view
  synthesis of a dynamic scene from monocular video.
\newblock In {\em Proceedings of the IEEE/CVF International Conference on
  Computer Vision (ICCV)}, pages 12959--12970, October 2021.

\bibitem{Wang2004image}
Zhou {Wang}, A.~C. {Bovik}, H.~R. {Sheikh}, and E.~P. {Simoncelli}.
\newblock Image quality assessment: from error visibility to structural
  similarity.
\newblock {\em TIP}, 13(4):600--612, 2004.

\bibitem{Wizadwongsa2021nex}
Suttisak Wizadwongsa, Pakkapon Phongthawee, Jiraphon Yenphraphai, and Supasorn
  Suwajanakorn.
\newblock Nex: Real-time view synthesis with neural basis expansion.
\newblock In {\em Proceedings of the IEEE/CVF Conference on Computer Vision and
  Pattern Recognition (CVPR)}, pages 8534--8543, June 2021.

\bibitem{Xian2021space}
Wenqi Xian, Jia-Bin Huang, Johannes Kopf, and Changil Kim.
\newblock Space-time neural irradiance fields for free-viewpoint video.
\newblock In {\em Proceedings of the IEEE/CVF Conference on Computer Vision and
  Pattern Recognition (CVPR)}, pages 9421--9431, June 2021.

\bibitem{Xue2019video}
Tianfan Xue, Baian Chen, Jiajun Wu, Donglai Wei, and William~T Freeman.
\newblock Video enhancement with task-oriented flow.
\newblock {\em International Journal of Computer Vision (IJCV)},
  127(8):1106--1125, 2019.

\bibitem{Zhang2021separable}
Feihu Zhang, Oliver~J. Woodford, Victor~Adrian Prisacariu, and Philip~H.S.
  Torr.
\newblock Separable flow: Learning motion cost volumes for optical flow
  estimation.
\newblock In {\em Proceedings of the IEEE/CVF International Conference on
  Computer Vision (ICCV)}, pages 10807--10817, 2021.

\bibitem{Zhang2018the}
Richard Zhang, Phillip Isola, Alexei~A. Efros, Eli Shechtman, and Oliver Wang.
\newblock The unreasonable effectiveness of deep features as a perceptual
  metric.
\newblock In {\em Proceedings of the IEEE Conference on Computer Vision and
  Pattern Recognition (CVPR)}, June 2018.

\bibitem{zhang2018perceptual}
Richard Zhang, Phillip Isola, Alexei~A Efros, Eli Shechtman, and Oliver Wang.
\newblock The unreasonable effectiveness of deep features as a perceptual
  metric.
\newblock In {\em CVPR}, pages 586--595, 2018.

\bibitem{Zhou2018stereo}
Tinghui Zhou, Richard Tucker, John Flynn, Graham Fyffe, and Noah Snavely.
\newblock Stereo magnification: Learning view synthesis using multiplane
  images.
\newblock {\em ACM Trans. Graph.}, 37(4), 2018.

\end{thebibliography}
}

\clearpage

\twocolumn[{%
\renewcommand\twocolumn[1][]{#1}%
% \maketitle
\begin{center}
    \centering
    \captionsetup{type=figure}
    \includegraphics[width=\textwidth]{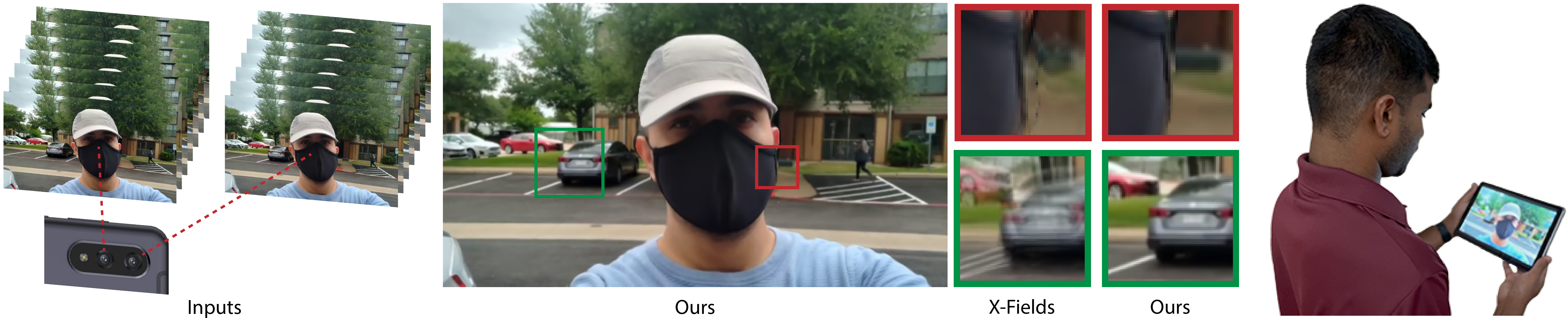}
    \captionof{figure}{We propose a lightweight approach to reconstruct images at novel views and times from a stereo video, captured with standard cameras (e.g., cellphones). We build upon X-Fields and propose several key ideas including regularization losses and non-linear coordinates to significantly improve the results. Our method runs in near real-time rates (23 fps) and has low memory and storage costs. Our system can be deployed on VR and light field displays to provide an immersive experience for the users (Lume Pad with a light field display shown on the right).}
    \label{fig:teaser}
\end{center}%
}]

\setcounter{section}{0}
\renewcommand{\thesection}{\Alph{section}} 
\begin{center}
    {\bf\LARGE - Supplementary -}
\end{center}

In this supplementary document, we provide additional implementation details, ablation results, and qualitative comparisons.

\section{Blending Network}
In this section, we provide further details about the blending network training. As discussed in the paper, we train a UNet on the Vimeo90K~\cite{Xue2019video} dataset to generate weight maps to blend the warped frames. Each scene in Vimeo90K consists of 7 frames. We use the corner frames (1 and 7) as reference frames and randomly select an intermediate frame from 2-6. We apply data augmentation techniques like random horizontal and vertical flip, and cropping to the sequence. We use Zhang et al.'s optical flow method~\cite{Zhang2021separable} to estimate the flow between intermediate and reference frames. The reference frames are warped using these flows and the weight maps from the UNet are used to smoothly blend these warped frames. We use the following loss to train our blending network:

\begin{equation}
    \mathcal{L} = \lambda_r \mathcal{L}_r + \lambda_p \mathcal{L}_p
\end{equation}

\noindent where $\lambda_r$ and $\lambda_p$ are set to 100 and 0.05 in our implementation. The two terms are described below.

\noindent\textbf{Reconstruction Loss $\mathcal{L}_r$}\quad We use the commonly used pixel-wise $\mathcal{L}_1$ loss~\cite{Niklaus2018context, Jiang2018super} between the blended image, $\tilde{I}_t$, and the ground truth intermediate frame, ${I}_t$.
\begin{equation}
    \mathcal{L}_r = \Vert \tilde{I}_t - I_t \Vert_1.
\end{equation}

\noindent\textbf{Perceptual Loss $\mathcal{L}_p$}\quad In addition, we utilize a VGG based perceptual loss~\cite{Jiang2018super, Niklaus2018context} to improve the details in the blended image \cite{zhang2018perceptual}. Specifically, we define the loss function as:
\begin{equation}
    \mathcal{L}_p = \Vert \phi(\tilde{I}_t) - \phi(I_t) \Vert_2^2,
\end{equation}
\noindent where $\phi$ is the response of \texttt{conv4\_3} layer of the pre-trained VGG-16 network.
\vspace{0.1in}

\begin{figure}
\includegraphics[width=\linewidth]{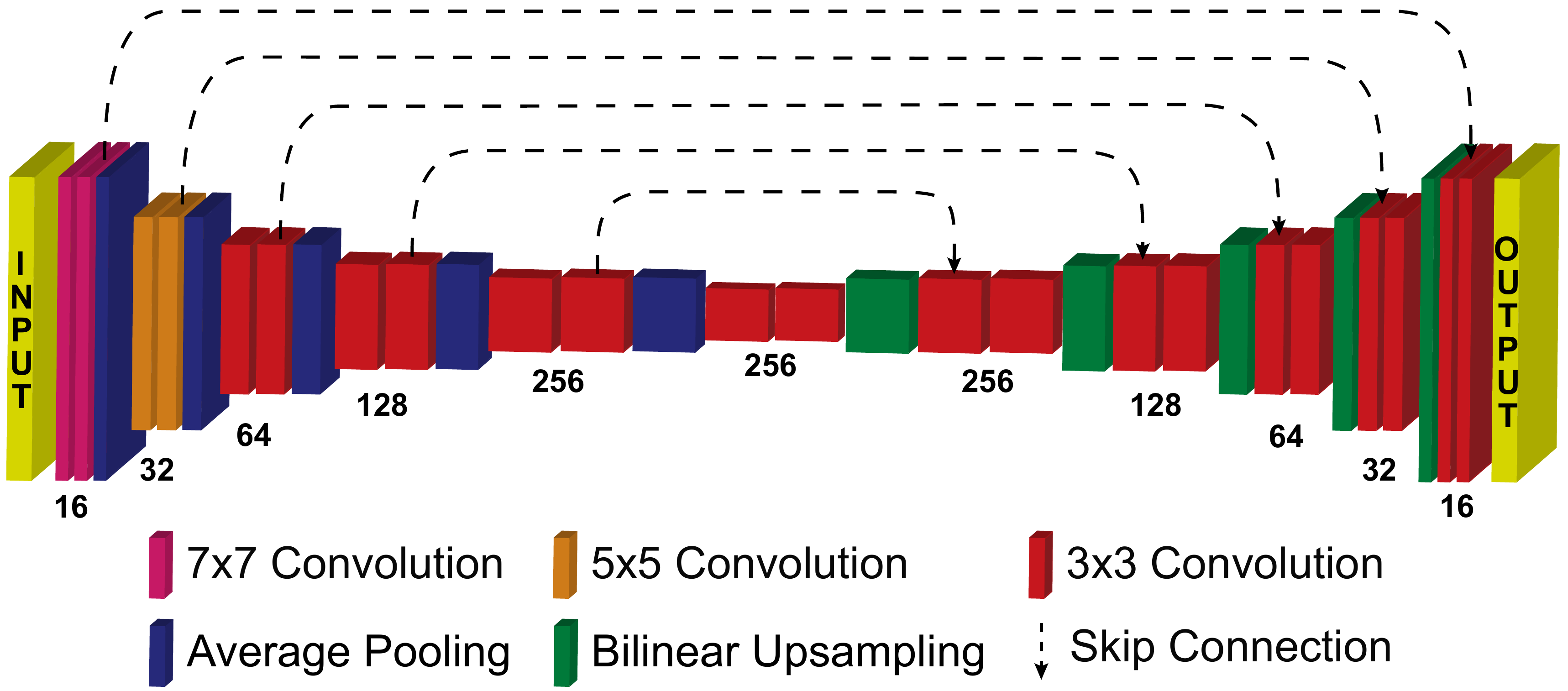}
\caption{This figure shows the UNet architecture used for our blending network. We use a LeakyReLU activation function for all the layers except the output for which we use sigmoid.}
\label{fig:arch}
\end{figure}

\noindent\textbf{Architecture}\quad The UNet architecture is described in Fig.~\ref{fig:arch}. This network takes a 16 channel input (two input images, warped images and the corresponding flows) and estimates a one channel weight map. We use a LeakyReLU activation function for all the layers except the output for which we use sigmoid.

\section{Multi-plane Disparity}
\begin{figure*}
\includegraphics[width=\linewidth]{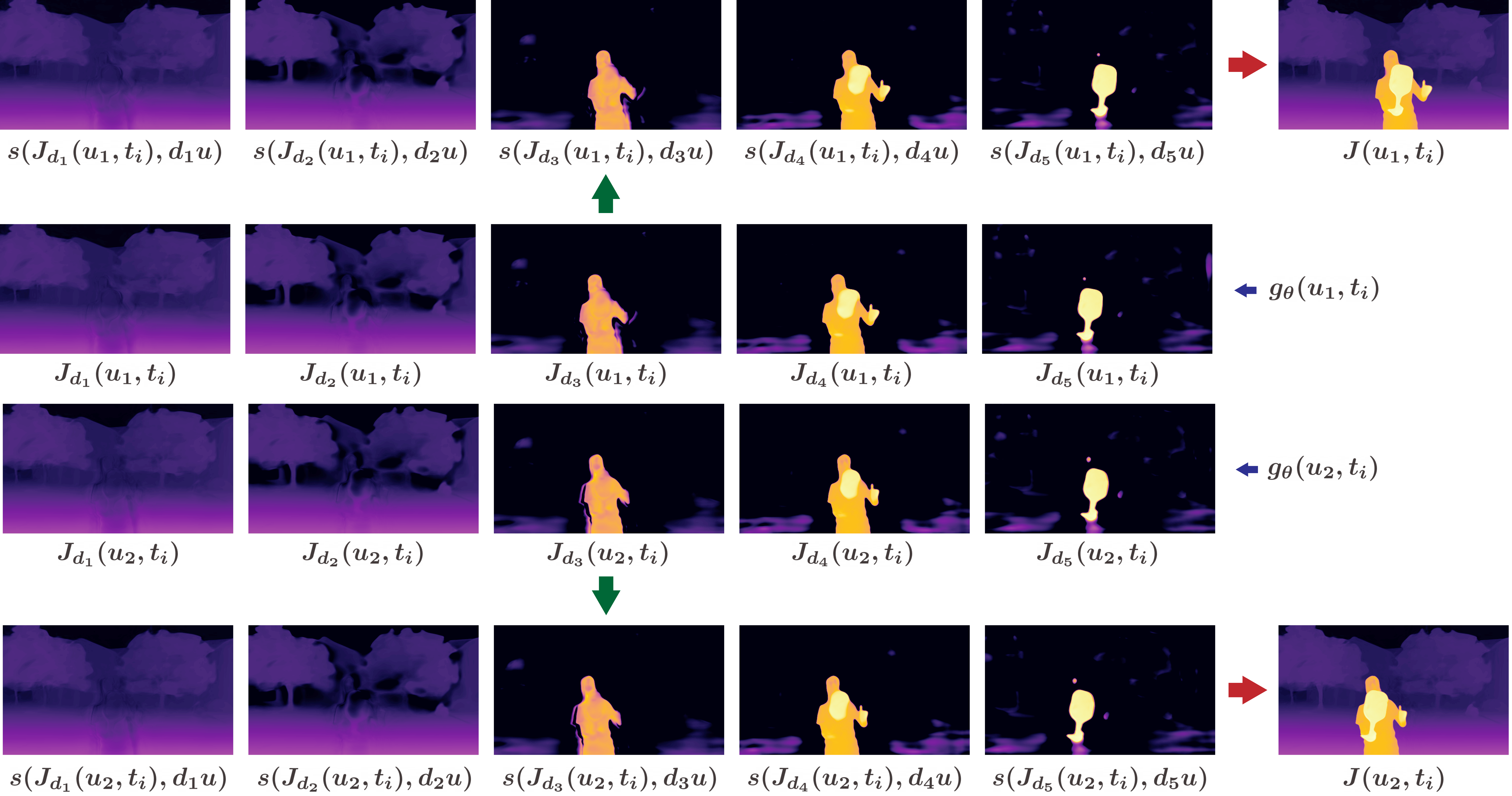}
\caption{This figure demonstrates the multi-plane disparity approach with a real example. The network uses 5 planes to encode the scene disparity. You can see that the network learns to separate objects ($g_\theta(u_j, t_i)$) into separate planes ($J_{d_k}(u_j, t_i)$) with similar texture. For both views $u_j$, the objects in each plane are close to each other which enables high quality interpolation. These planes are then shifted ($s(J_{d_k}(u_j, t_i), d_ku)$) and merged using pixel-wise \texttt{max} operation to obtain the final view Jacobian ($J(u_j, t_i)$).}
\label{fig:shift}
\end{figure*}

We provide another illustrative figure (Fig.~\ref{fig:shift}) for multi-plane disparity using a real example. The network outputs multi-planes with objects close to each other in both views $u_j$. These planes are shifted and merged using the \texttt{max} operator to obtain the final disparity, $J(u_j, t_i)$. Although the reference view Jacobian has large disparity, the encoded objects are spatially close which enables high quality interpolation.

\section{Quantitative Results Dataset}
In this section, we describe the steps to prepare the quantiative evaluation dataset used in our paper. We numerically compare our approach against the other methods on two lightfield video datasets, Sintel~\cite{Kinoshita2021depth} and LFVID~\cite{Sabater2017dataset}. Sintel dataset contains 23 9x9 lightfield videos of 7 distinct scenes each with 20 to 50 frames. Among these we selected 5 sequences from distinct scenes that had larger number of frames (2 scenes contained only short sequences): \texttt{ambushfight\_5}, \texttt{bamboo\_1}, \texttt{chickenrun\_3}, \texttt{foggyrocks\_1}, as well as \texttt{shaman\_2}. We use every other frame of the sequences \texttt{05\_00} and \texttt{05\_08} as the input left and right videos and the sequence \texttt{05\_04} as our ground truth. We center crop the images to 360p. The LFVID dataset contains $4\times4$ light field videos, from which, we select 6 scenes: \texttt{Birthday}, \texttt{Hands}, \texttt{Painter}, \texttt{Rugby}, \texttt{Theater} and \texttt{Train}. We use every other frame of the videos from views 0 and 2 (from the second row) as our input and the video for view 1 as the ground truth. We downsample all the images by a factor of 2 and center crop them to 360p.

\begin{table}[!t]
\renewcommand{\arraystretch}{1.3}
\caption{Effect of core components on view synthesis.}
\centering
\begin{footnotesize}
\begin{tabular}{l|ccc}
  \hline\hline
  & PSNR & SSIM & LPIPS\\
  \hline
  Appearance Loss & 27.11 & 0.843 & 0.0519\\
  \quad+ Mask + Jacobian Loss & 28.34 & 0.870 & 0.0460\\
  \quad + Positional Encoding & 28.49 & 0.868 & 0.0456\\
  \quad + Multi-plane disparities & \textbf{33.39} & \textbf{0.935} & \textbf{0.0242}\\
  \hline\hline
\end{tabular}
\label{tab:ablationview}
\end{footnotesize}
\end{table}

\begin{table}[!t]
\renewcommand{\arraystretch}{1.3}
\caption{Effect of components on multi-plane disparity training.}
\centering
\begin{footnotesize}
\begin{tabular}{l|ccc}
  \hline\hline
  & PSNR & SSIM & LPIPS\\
  \hline
  w/o per plane regularization & 28.41 & 0.892 & 0.0428\\
  w/o disparity mask & 31.48 & 0.924 & 0.0322\\
  w/ pixel-wise sum & 31.51 & 0.907 & 0.0316\\
  w/ pixel-wise max (Ours) & \textbf{33.39} & \textbf{0.935} & \textbf{0.0242}\\
  \hline\hline
\end{tabular}
\label{tab:ablationmultiplane}
\end{footnotesize}
\end{table}

\begin{table}[!t]
\renewcommand{\arraystretch}{1.3}
\caption{Effect of number of planes on view synthesis.}
\centering
\begin{footnotesize}
\begin{tabular}{l|ccc}
  \hline\hline
  \#planes & PSNR & SSIM & LPIPS\\
  \hline
  2 & 27.66 & 0.860 & 0.0437\\
  3 & 30.72 & 0.927 & 0.0297\\
  4 & 32.13 & 0.930 & 0.0266\\
  6 & \textbf{33.39} & 0.935 & 0.0242\\
  9 & 33.23 & \textbf{0.937} & \textbf{0.0235}\\
  \hline\hline
\end{tabular}
\label{tab:ablationnumplanes}
\end{footnotesize}
\end{table}

% \begin{table}[!t]
% \renewcommand{\arraystretch}{1.3}
% \caption{Effect of components on time interpolation.}
% \centering
% \begin{footnotesize}
% \begin{tabular}{l|ccc}
%   \hline\hline
%   & PSNR & SSIM & LPIPS\\
%   \hline
%   Single Jacobian & 39.10 & 0.963 & 0.0115\\
%   Dual Jacobian & 39.13 & 0.963 & 0.0114\\
%   Non-Uniform coordinates & \textbf{39.45} & \textbf{0.966} & \textbf{0.0108}\\
%   \hline\hline
% \end{tabular}
% \label{tab:ablationtime}
% \end{footnotesize}
% \end{table}

\begin{table}[!t]
\renewcommand{\arraystretch}{1.3}
\caption{Effect of components on time interpolation.}
\centering
\begin{footnotesize}
\begin{tabular}{l|ccc}
  \hline\hline
  & PSNR & SSIM & LPIPS\\
  \hline
  Single Jacobian & 32.88 & 0.954 & 0.0261\\
  Dual Jacobian & 33.12 & 0.954 & 0.0259\\
  Non-Uniform coordinates & \textbf{33.59} & \textbf{0.959} & \textbf{0.0235}\\
  \hline\hline
\end{tabular}
\label{tab:ablationtime}
\end{footnotesize}
\end{table}

\section{Ablation Experiments}
Here, we evaluate the effectiveness of various components in our model. We use the \texttt{Painter} scene from LFVID~\cite{Sabater2017dataset} dataset for numerical ablations of view synthesis. In Table~\ref{tab:ablationview}, we show the effectiveness of core components of view synthesis algorithm. The Jacobian loss with masked appearance loss significantly improves the output quality by applying occlusion aware training. Positional encoding improves the fine details of encoded Jacobians. Our multi-plane approach drastically improves the output quality as it is able to handle scenes with large disparity.

\begin{figure*}
\includegraphics[width=\linewidth]{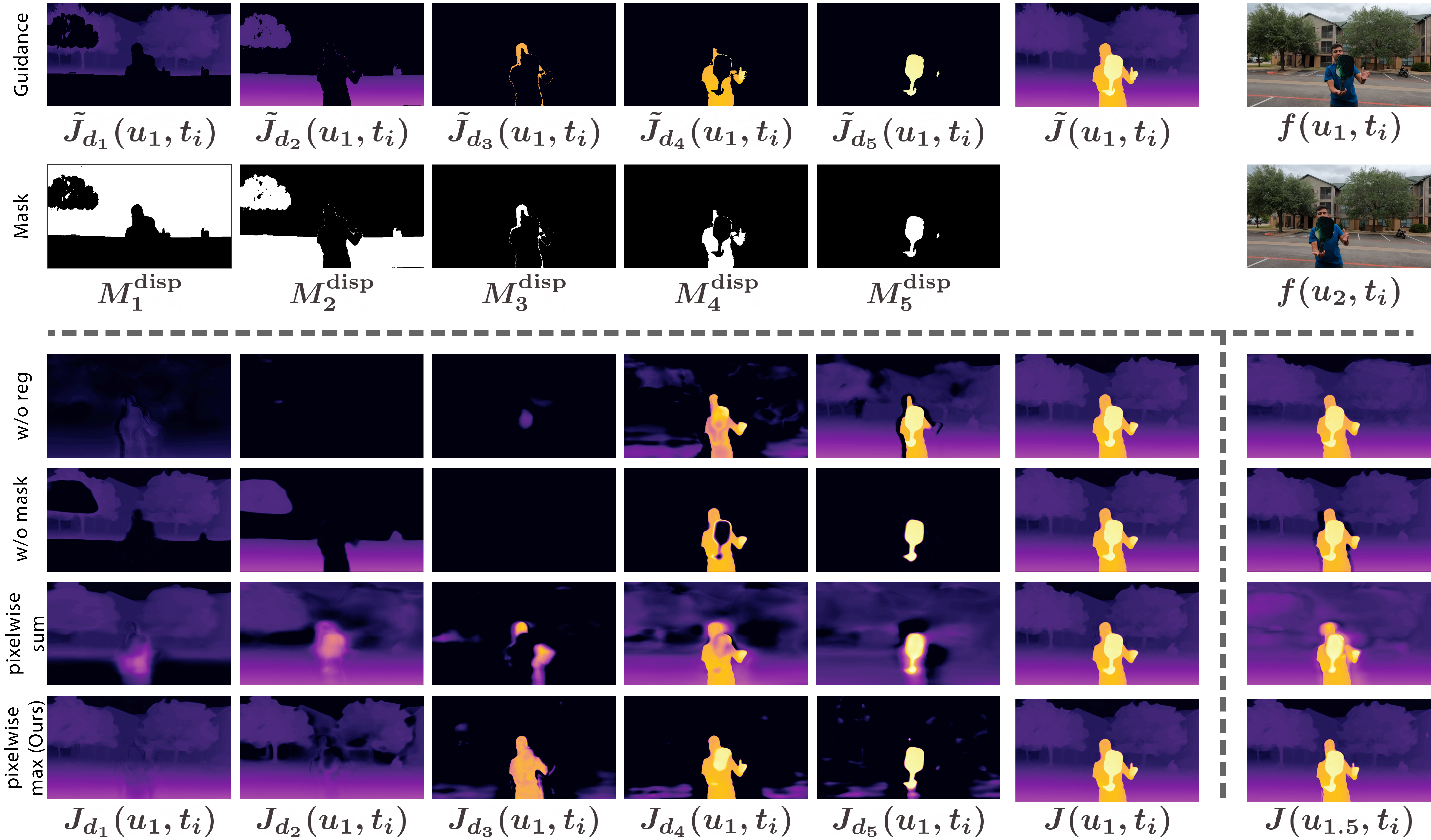}
\caption{The top two rows consist of per plane guidance Jacobians ($\tilde{J}_{d_k}(u_1, t_i)$), disparity masks ($M_{k}^{disp}$), guidance Jacobian ($\tilde{J}(u_1, t_i)$) and the input reference frames ($f(u_j, t_i)$). The bottom four rows qualitatively demonstrate the effect of different components on multi-plane disparity training. Note that all configurations are able to encode guidance Jacobians correctly ($J(u_1, t_i)$). However, the quality of intermediate Jacobian used for view synthesis ($J(u_{1.5}, t_i)$) changes depending on the setting. Without per plane regularization (first row), the network tends to encode all information in a single plane ($J_{d_5}(u_1, t_i)$) which leads to loss of details in the intermediate Jacobian. Without disparity mask (second row), the network has a hard constraint to exactly encode the per plane guidance Jacobians which leads to artifacts around plane boundaries during view synthesis. Using \texttt{sum} operator to merge planes generates severe artifacts (third row) at intermediate coordinates. Our approach with \texttt{max} operator (fourth row) is able to generate high quality intermediate Jacbians.}
\label{fig:multiplane_supp}
\end{figure*}

We further explore the components of multi-plane disparity approach in Table~\ref{tab:ablationmultiplane}. Without the per plane regularization, we see a drastic drop in performance highlighting its importance. Training without the disparity mask negatively affects the performance and creates sharp artifacts in the output. Using the pixel-wise \texttt{max} operator over \texttt{sum} to merge the planes gives a significant performance boost to our model. In addition, we demonstrate the effect of these components visually in Fig.~\ref{fig:multiplane_supp}. The first two rows show the guidance Jacobians and disparity masks for per plane regularization. The bottom four rows show the effect of multi-plane disparity components. While all the configurations are able to encode the guidance Jacobians ($J(u_1, t_i)$), the intermediate Jacobian ($J(u_{1.5}, t_i)$) contains artifacts for all the other configurations. Without per plane regularization, the network tends to put most of the details in one plane ($J_{d_5}(u_1, t_i)$). Note that while a couple of other planes contain content, they will be masked by the the fifth plane as it contains maximum disparity in most pixels. This leads to missing details in the intermediate Jacobian. Without disparity mask, the network has no flexibility in encoding the objects at different disparity. This ends up creating artifacts around plane boundaries (halo around the person) in the intermediate Jacobian. Using the pixel-wise \texttt{sum} operator creates fuzzy intermediate Jacobian. Our configuration with pixel-wise \texttt{max} operator creates detailed intermediate Jacobian without any significant boundary artifacts.

In Table~\ref{tab:ablationnumplanes}, we show the effect of number of planes on the output quality. While the performance keeps improving until 6 planes, we do not observe significant improvement in performance beyond that. So, we choose $\#planes=6$ to maximize the view synthesis quality while limiting the computation cost.

\begin{figure}
\includegraphics[width=\linewidth]{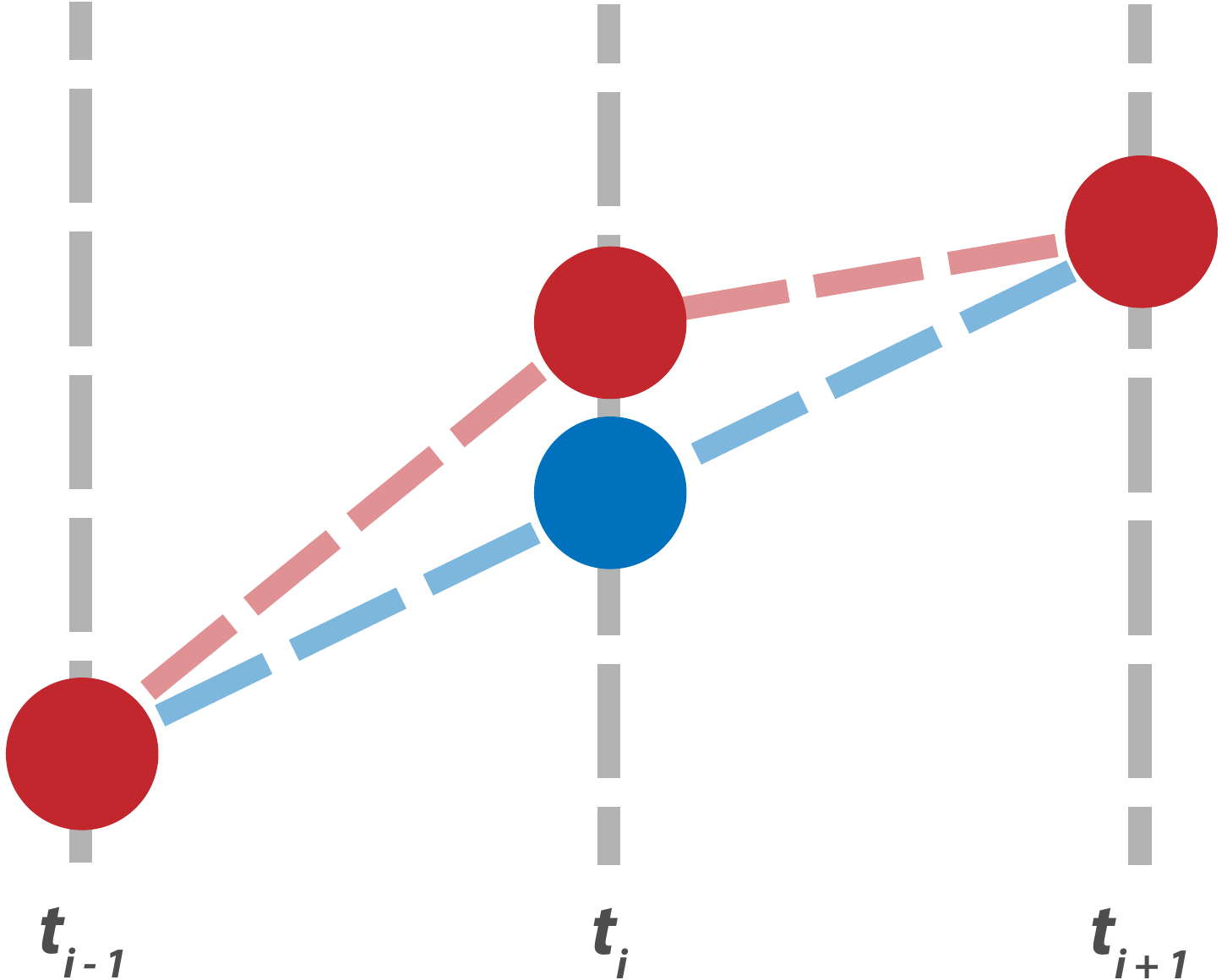}
\caption{Here, we demonstrate the non-linear motion that frequently occurs in videos. The red dots show actual pixel motion at different time steps. Using a single Jacobian value to encode pixel motion to both left and right time steps results in averaging (blue dot) which leads to ghosting artifacts.}
\label{fig:nonlinear_supp}
\end{figure}

We use the \textsc{Cars} scene from our GoPro rig for numerical ablations of time interpolation. Since we captured the GoPro scenes at 240fps, we have ground truth frames for this experiment. In Table~\ref{tab:ablationtime}, we look at the time interpolation components of our model. Single Jacobian is unable to handle non-linear motion, as illustrated in Fig.~\ref{fig:nonlinear_supp}. Using dual Jacobian is able to improve the time interpolation quality by handling small non-linear motion. Using non-uniform coordinates improves the performance by a good margin as it allows the network to encode highly non-linear motion and motion spikes. The effectiveness of non-uniform coordinates becomes apparent in visual comparisons. E.g., every time there is sudden change in camera or object motion in the supplementary video, X-Fields will generate glaring artifacts while our approach is able to smoothly handle them.
% \begin{figure*}
% \includegraphics[width=\linewidth]{Images/frameinterpcomparison.pdf}
% \caption{Time-interpolation comparison against the state-of-the-art frame interpolation approach by Niklaus et al.~\cite{Niklaus2021revisiting} (SepConv++). As seen the left and right interpolated frames by SepConv++ contain significant artifacts and are not consistent. Our approach, however, produces significantly better results.}
% \label{fig:frameinterpcomparison}
% \end{figure*}

\begin{figure*}
\includegraphics[width=\linewidth]{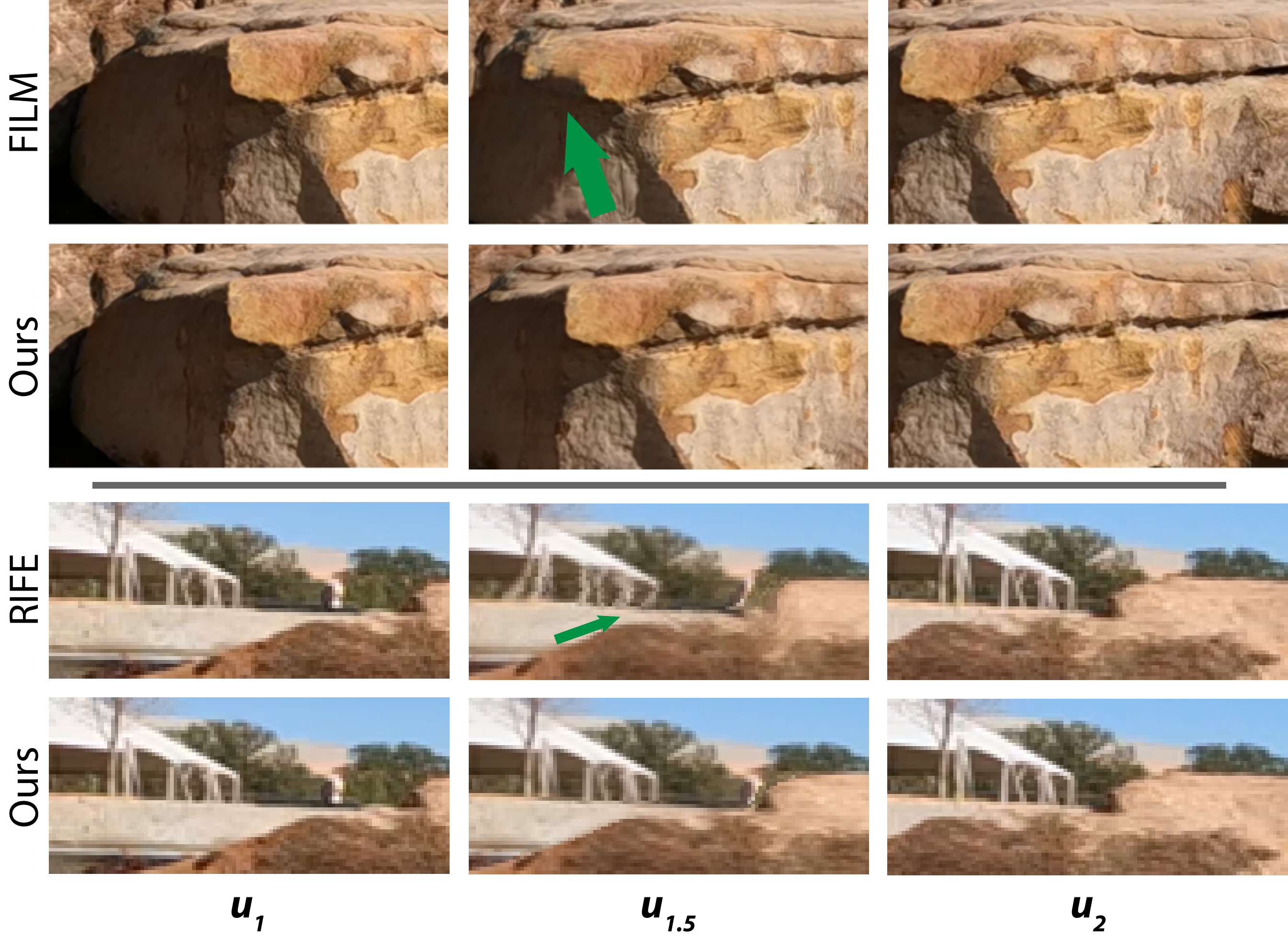}
\caption{Another approach to view-time synthesis would be to use a video interpolation model to perform both view and time synthesis. However, without the constraint of static motion, these models tend to introduce artifacts that result in temporal inconsistency. E.g., while FILM~\cite{reda2022film} is able to handle large motion, it sometimes deforms the texture to smoothly animate between the two views. Our approach is able to correctly warp the texture. RIFE~\cite{huang2022rife} on the other hand, while able to synthesize frames in real-time, struggles at motion boundaries.}
\label{fig:filmviewsyn}
\end{figure*}

\section{Results}
\label{sec:Results_supp}

\begin{figure}
\includegraphics[width=\linewidth]{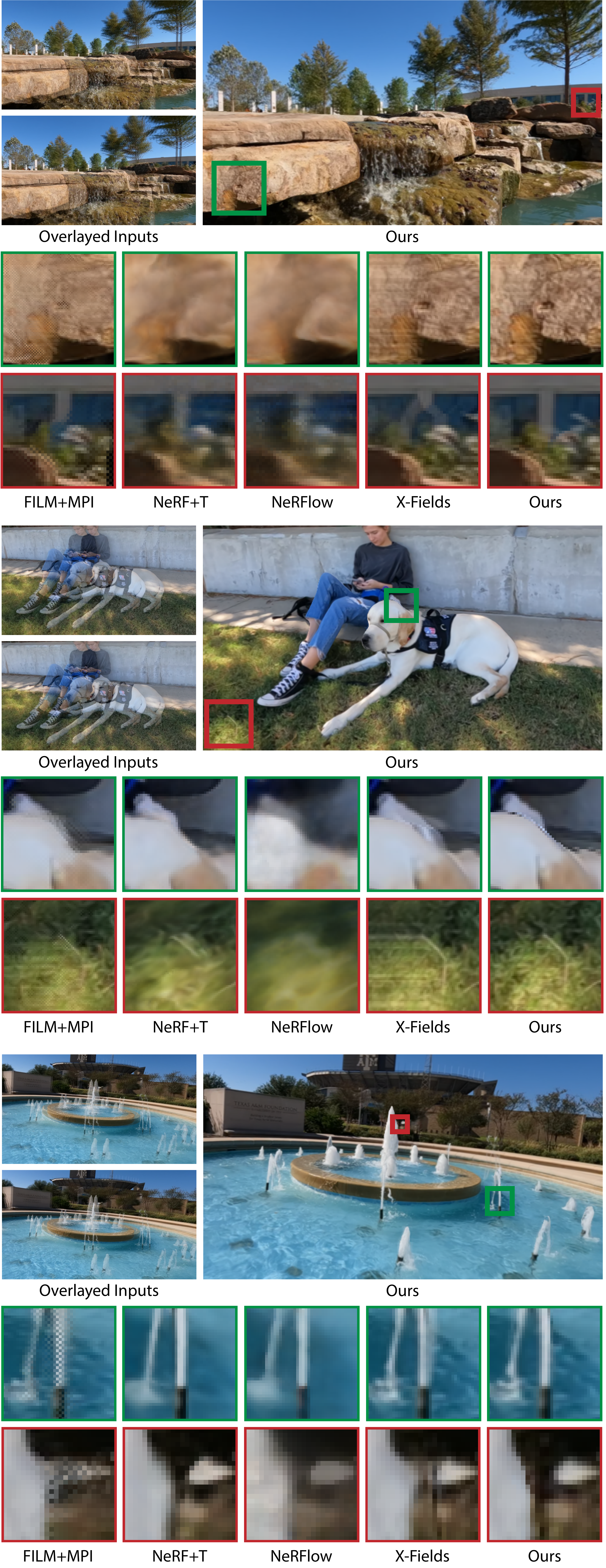}
\vspace{-0.2in}
\caption{Comparison against several state-of-the-art methods on view-time interpolation. On the left we show the overlayed left and right views for two consecutive frames neighboring the coordinate of interest.}
\label{fig:viewtime_results1}
\vspace{-0.2in}
\end{figure}

\begin{figure}
\includegraphics[width=\linewidth]{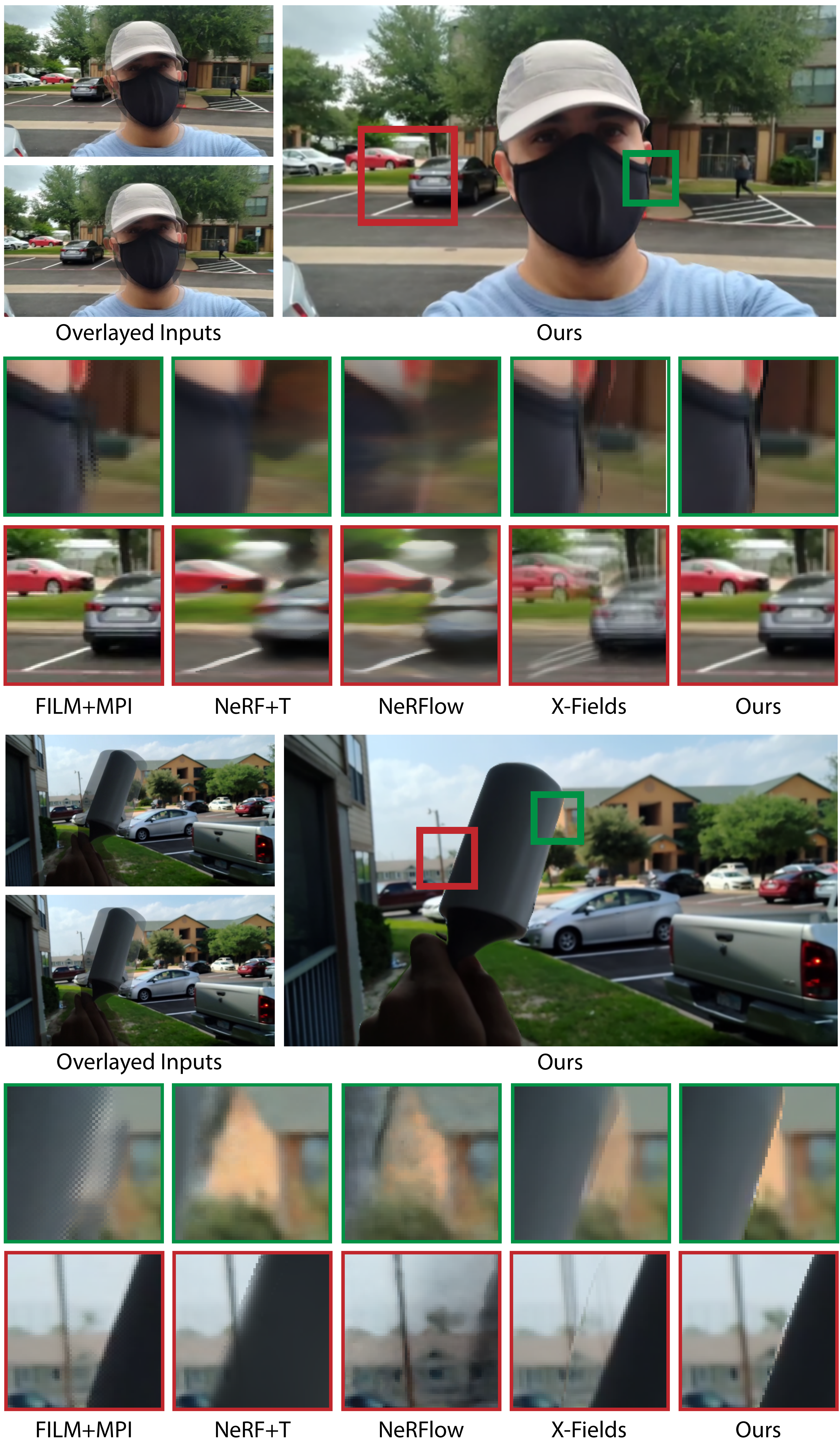}
\vspace{-0.2in}
\caption{Comparison against several state-of-the-art view-time interpolation methods on Lume Pad~\cite{leia} scenes. On the left we show the overlayed left and right views for two consecutive frames neighboring the coordinate of interest.}
\label{fig:viewtime_results3}
\vspace{-0.2in}
\end{figure}

\subsection{Qualitative Results}
\begin{figure*}
\includegraphics[width=\linewidth]{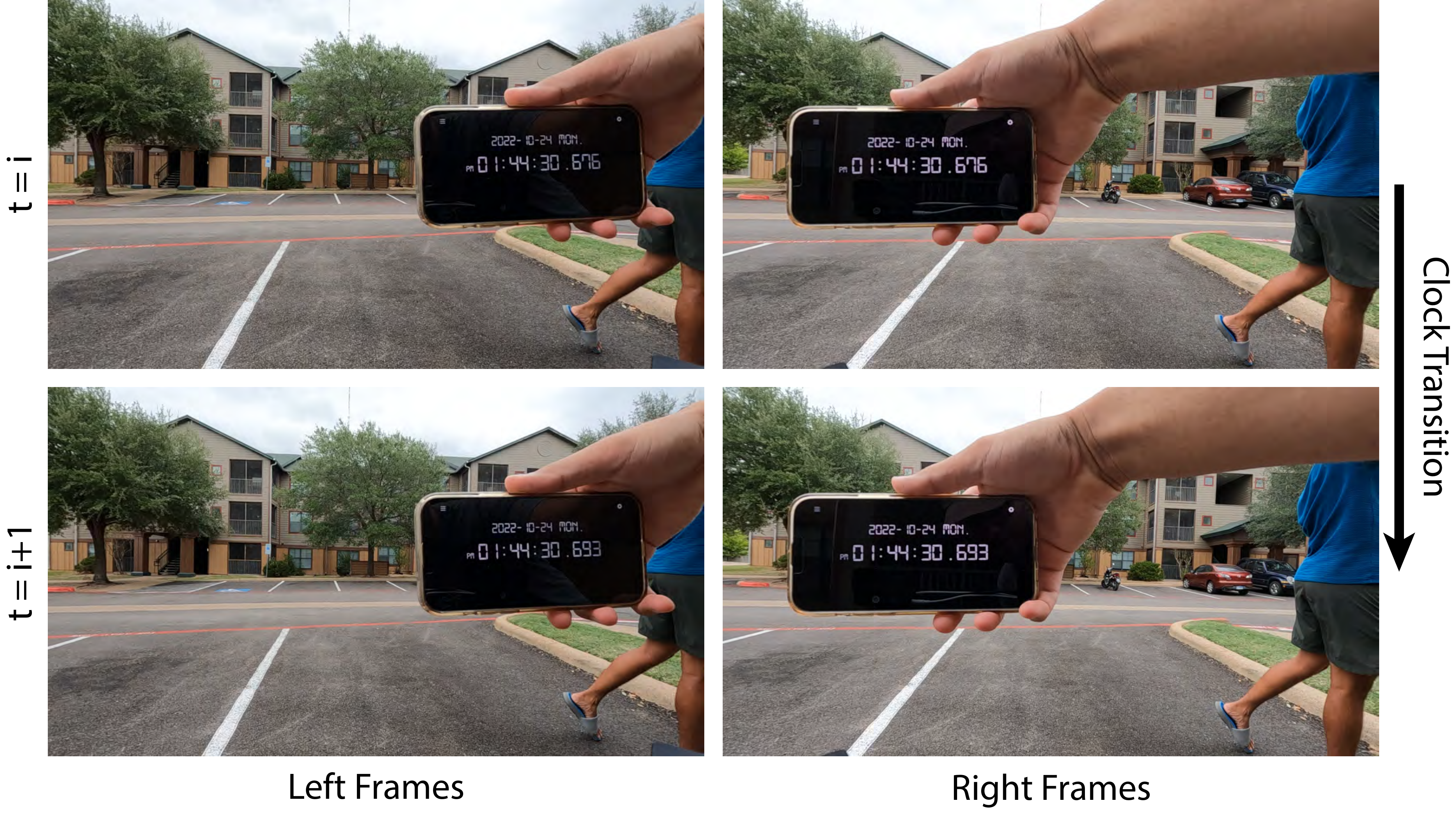}
\vspace{-0.2in}
\caption{We observe the frames where the clock transitions to the next time stamp in both left and right frames. This transition is used to synchronize the stereo videos to the closest frame.}
\label{fig:timesync}
\vspace{-0.2in}
\end{figure*}

We capture a set of stereo videos with a variety of motions using a stereo GoPro camera rig and Lume Pad~\cite{leia} (stereo rear cameras as shown in Fig.~\ref{fig:teaser}). The Lume Pad videos are automatically time synchronized and rectified. For the GoPro stereo videos, we time synchronize the left and right videos using a high precision clock and synchronizing the clock transitions as shown in Fig.~\ref{fig:timesync}. We then perform uncalibrated stereo video rectification on the synchronized stereo videos.

We show comparisons against several state-of-the art approaches. Fig.~\ref{fig:viewtime_results1} shows videos captured using the GoPro rig. The Lume Pad videos are shown in Fig.~\ref{fig:viewtime_results3}. For all the scenes, we show view-time interpolation at the middle of four observed view-time frames. As seen, other approaches produce results with noticeable ghosting and other artifacts, while our results are sharp and have clear boundaries.

We can also use video interpolation approaches like FILM~\cite{reda2022film} and RIFE~\cite{huang2022rife} for view synthesis. However, these approaches tend to generate artifacts, as shown in Fig~\ref{fig:filmviewsyn}, which leads to temporal inconsistency.

\end{document}